\documentclass[10pt,twocolumn,letterpaper]{article}
\usepackage[pagenumbers]{cvpr}  %

\usepackage{currfile} %

\usepackage{caption} %

\newcommand{\ms}[2]{#1{\scriptsize$\pm$#2}}

\newcommand{\natexlab}[1]{}%

\makeatletter
\renewcommand\paragraph{\@startsection{paragraph}{4}{\z@}{1.2ex \@plus .3ex \@minus .2ex}{-1em}{\normalfont\normalsize\bfseries}}
\makeatother
\AtBeginDocument{\crefname{table}{Table}{Tables}}%
\usepackage{booktabs}
\usepackage{multirow}
\usepackage{amsmath,amssymb}
\usepackage{iftex}
\ifXeTeX
  \usepackage[no-math]{fontspec}
\fi
\usepackage{xcolor}
\usepackage{algorithm}
\usepackage{algpseudocode}
\definecolor{cvprblue}{rgb}{0.21,0.49,0.74}
\usepackage[pagebackref,breaklinks,colorlinks,allcolors=cvprblue]{hyperref}

\def\confName{CVPR}
\def\confYear{2027}

\title{Change Partners at Every Layer:\\ Balanced Random Regrouping for Model Merging}

\author{Kaiyang Li$^{1}$\quad Shaobo Han$^{2}$\quad Shihao Ji$^{1}$\\
$^{1}$University of Connecticut\quad $^{2}$NEC Laboratories America\\
{\tt\small \{kaiyang.li, shihao.ji\}@uconn.edu, shaobo@nec-labs.com}}

\begin{document}
\maketitle
\begin{abstract}
Model merging combines fine-tuned experts into one model that serves many tasks without retraining, but merging many experts this way substantially reduces accuracy.
Grouping offers a middle ground: at every layer or weight matrix, the $N$ experts are divided into $K<N$ groups, the experts in a group are merged into one shared module, and a task uses the module merged from its expert's group.
Natural choices are to keep the same partners across layers or to group similar experts.
Surprisingly, we find that randomly changing partners at every layer outperforms both.
We call this Balanced Random Regrouping (BRR): it independently partitions the experts into $K$ equal-sized random groups at each layer (BRR-layer) or weight matrix (BRR-matrix) and merges each group with an existing method.
With validation-selected merge scales, both BRR variants outperform reusing one grouping across layers in all 24 CLIP settings, by 1.0--6.8 points, and BRR-matrix also outperforms the similarity-based FlexMerge in every setting.
A first-order analysis explains this gain.
At each layer, merging a task's expert with its partners shifts the task's output features; when the partners are the same at every layer, these shifts tend to point in similar directions and add up, whereas with new partners at every layer they vary independently and add up less on average.
A series of experiments supports this analysis, which further guides a grouping search that can outperform BRR.

\end{abstract}

\begin{figure*}[t]
  \centering
  \includegraphics[width=\textwidth]{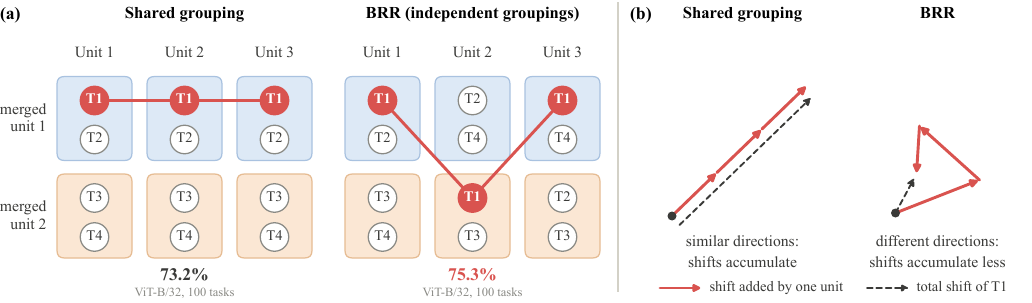}
  \caption{\textbf{Shared grouping versus BRR (schematic).} (a)~Four tasks (T1--T4) and three units (layers or weight matrices); at each unit, the experts are merged in two equal-sized groups. Shared grouping reuses one grouping, whereas BRR draws one independently at each unit; the red line follows the modules selected by T1. The accuracies shown are those of shared grouping and BRR-layer on 100 ViT-B/32 tasks (TSV-M, $K{=}10$, $\alpha{=}1$). (b)~Under a first-order approximation, each red arrow shows the shift that merging at one unit adds to T1's final output features, and the dashed arrow is their sum. In this illustration, fixed partners give shifts in similar directions that accumulate, whereas new partners give shifts in different directions that accumulate less.}
  \label{fig:method}
  \vspace{-8pt}
\end{figure*}

\section{Introduction}\label{sec:intro}

Model merging combines fine-tuned experts into one model that serves many tasks without retraining~\cite{ilharco2023ta,yadav2023ties,gargiulo2025tsv}, but accuracy drops substantially when many experts are merged.
Keeping every expert avoids this loss but costs one full model per task, whereas many deployments can store only a few models~\cite{dhasade2026flexmerge}.
Merging the experts in groups lies between these two extremes: at each layer, the $N$ experts are divided into $K<N$ groups, and each group is merged into one shared module.
At inference, the task identity is known, so each input is routed at every layer to the module merged from its task's expert group.
The grouping can differ from layer to layer.
The core question is therefore which experts to merge together at each layer.
\par

Two design choices seem natural.
One reuses a single grouping across layers, which yields $K$ merged models with each expert keeping the same partners at every layer.
The other groups experts by parameter similarity, as FlexMerge~\cite{dhasade2026flexmerge} does for attention and MLP blocks and Channel Merging~\cite{zhang2025channel} does for rows of weight matrices.
Surprisingly, we find that randomly changing partners at every layer outperforms both choices.
We call this method \emph{Balanced Random Regrouping} (BRR).
Specifically, at each transformer layer, BRR randomly divides the experts into $K$ balanced (equal-sized) groups, independently of other layers (Fig.~\ref{fig:method}(a)).
The corresponding parameters within each group are merged with an existing method, such as Task Arithmetic~\cite{ilharco2023ta}, TSV-M~\cite{gargiulo2025tsv}, Iso-CTS~\cite{marczak2025iso}, or ACE-Merging~\cite{xu2026ace}.
We further extend the grouping from layers to weight matrices, allowing partners to change at a finer granularity; we call the two variants BRR-layer and BRR-matrix.
\par

We evaluate BRR on 100-task CLIP~\cite{radford2021clip} expert pools with two backbones, four merging methods, and three budgets.
With merge scales selected on validation data for each method, both BRR variants outperform shared grouping (one random grouping reused across layers) in all 24 configurations, and BRR-matrix also achieves higher mean accuracy than the similarity-based FlexMerge~\cite{dhasade2026flexmerge} in every configuration.
On two vision-language models, BRR-matrix outperforms both baselines with all four merging methods on InternVL2.5-1B~\cite{chen2024internvl25}, and improves on shared grouping by 0.60 points on average on Qwen2-VL-7B~\cite{wang2024qwen2vl}.
\par

To understand these gains, we analyze how merging at different layers jointly changes a task's final output features.
Intuitively (Fig.~\ref{fig:method}(b)), merging at each layer shifts the final output features; with fixed partners, these shifts tend to reinforce one another, whereas changing partners reduces this accumulation.
Our first-order analysis makes this precise: reusing partners adds a correlation between the shifts of different layers, and regrouping independently at each layer removes it, which lowers the expected feature error to first order whenever this correlation is positive overall.
Under Task Arithmetic on ViT-B/32, our estimates of this correlation are positive for each of the 100 tasks.
The analysis also lets us predict the accuracy of merged models from those of shared grouping and BRR.
For 27 groupings that reuse partners at only some units, whose gains over shared grouping range from 0.6 to 4.3 points, predictions fixed before their accuracies were available are off by only 0.16 points on average.
It further guides a grouping search that uses a few unlabeled validation images to choose partners and can outperform BRR.

Our contributions are summarized as follows:
\begin{itemize}
\item We show that randomly changing partners at every layer outperforms both natural choices, keeping the same partners and grouping similar experts, and propose BRR, which draws balanced random groups independently at each layer or weight matrix; its grouping step needs no data or search and works with existing merging methods.
\item On 100-task CLIP pools, both BRR variants outperform shared grouping in all 24 configurations, and BRR-matrix also outperforms FlexMerge in every configuration; on InternVL2.5-1B, BRR-matrix outperforms both shared grouping and FlexMerge with all four merging methods.
\item A first-order analysis explains why changing partners lowers feature error; a series of experiments supports this analysis, which further guides a grouping search that can outperform BRR.
\end{itemize}

\section{Related Work}\label{sec:related}

\textbf{Model merging.} Task Arithmetic~\cite{ilharco2023ta} merges experts by adding their task vectors, the differences between fine-tuned and pre-trained weights.
Later work proposes different ways to reduce the resulting interference.
Some methods modify task vectors in weight space: TIES~\cite{yadav2023ties} trims small updates and resolves sign conflicts, while TSV-M~\cite{gargiulo2025tsv} and Iso-CTS~\cite{marczak2025iso} decompose each layer's task updates with SVD and reduce interference between the resulting task subspaces.
Others use input statistics: RegMean~\cite{jin2023regmean} and ACE-Merging~\cite{xu2026ace} use measured or estimated statistics to keep each layer's output close to that of the experts.
EMR-Merging~\cite{huang2024emr} and TALL masks~\cite{wang2024tall} instead recover task-specific models with per-task masks or scaling factors.
BRR is complementary to these methods: it only decides which experts are merged together at each layer, and any of them can be applied within each group.
\par

\textbf{Merging experts in groups.} Some works merge experts into several models or shared modules instead of one, using more storage than a single merged model to retain task performance.
FlexMerge~\cite{dhasade2026flexmerge} groups network blocks by similarity, and Channel Merging~\cite{zhang2025channel} groups task vectors at the level of output channels.
For LoRA adapters, $D^2C$~\cite{bohdal2026d2c} uses a small set of task examples to optimize the clustering before merging each group, and K-Merge~\cite{shenaj2026kmerge} merges newly arriving LoRAs online without data under a fixed adapter budget.
In contrast, BRR forms its groups at random, without any similarity or clustering criterion, and asks a different question: whether an expert should keep the same partners across layers or change them.
\par

\textbf{Analyzing merging errors.} Task Arithmetic in the Tangent Space~\cite{ortiz2023tangent} studies task arithmetic through linearization and connects it to weight disentanglement.
Zhou et al.~\cite{zhou2024crosstask} provide a related empirical observation: interpolating weights fine-tuned from a common checkpoint approximately interpolates layer-wise features.
SurgeryV2~\cite{yang2024surgeryv2} observes that the representation bias introduced by merging at each layer accumulates through the network and corrects it with trained modules.
We instead use a first-order approximation to write the output-feature change as a sum of per-layer contributions, which shows that reusing partners adds a correlation between these contributions and that independent regrouping removes this added correlation.

\setcounter{dbltopnumber}{1}
\AddToHook{shipout/after}[brr-main-tables]{%
  \ifnum\value{page}=5
    \setcounter{dbltopnumber}{2}%
    \RemoveFromHook{shipout/after}[brr-main-tables]%
  \fi
}
\begin{table*}[t]
\centering\small
\setlength{\tabcolsep}{1.5pt}
\begin{tabular*}{\textwidth}{@{\extracolsep{\fill}}cl*{8}{c}@{}}
\toprule
& & \multicolumn{4}{c}{\textbf{ViT-B/32}} & \multicolumn{4}{c}{\textbf{ViT-L/14}} \\
\cmidrule(lr){3-6}\cmidrule(lr){7-10}
$K$ & Method & \makebox[41pt][c]{TA} & \makebox[41pt][c]{TSV-M} & \makebox[41pt][c]{Iso-CTS} & \makebox[41pt][c]{ACE} & \makebox[41pt][c]{TA} & \makebox[41pt][c]{TSV-M} & \makebox[41pt][c]{Iso-CTS} & \makebox[41pt][c]{ACE} \\
\midrule
& Zero-shot & \multicolumn{4}{c}{42.05} & \multicolumn{4}{c}{48.11} \\
$N$ & Experts & \multicolumn{4}{c}{85.68} & \multicolumn{4}{c}{89.79} \\
\midrule
1 & One model & \makebox[41pt][c]{43.26} & \makebox[41pt][c]{53.90} & \makebox[41pt][c]{54.58} & \makebox[41pt][c]{57.22} & \makebox[41pt][c]{48.81} & \makebox[41pt][c]{64.74} & \makebox[41pt][c]{66.74} & \makebox[41pt][c]{68.36} \\
\midrule
 & Channel Merging & \multicolumn{4}{c}{\makebox[41pt][c]{$\mathbf{53.55}${\scriptsize$\pm0.11$}}} & \multicolumn{4}{c}{\makebox[41pt][c]{$\mathbf{64.07}${\scriptsize$\pm0.07$}}} \\
 & FlexMerge & \makebox[41pt][c]{51.18} & \makebox[41pt][c]{66.90} & \makebox[41pt][c]{66.13} & \makebox[41pt][c]{70.36} & \makebox[41pt][c]{60.46} & \makebox[41pt][c]{76.13} & \makebox[41pt][c]{77.41} & \makebox[41pt][c]{79.36} \\
5 & Shared & \makebox[41pt][c]{$50.31${\scriptsize$\pm0.40$}} & \makebox[41pt][c]{$68.32${\scriptsize$\pm0.19$}} & \makebox[41pt][c]{$68.39${\scriptsize$\pm0.31$}} & \makebox[41pt][c]{$72.65${\scriptsize$\pm0.21$}} & \makebox[41pt][c]{$62.25${\scriptsize$\pm0.69$}} & \makebox[41pt][c]{$78.25${\scriptsize$\pm0.51$}} & \makebox[41pt][c]{$80.83${\scriptsize$\pm0.36$}} & \makebox[41pt][c]{$82.03${\scriptsize$\pm0.53$}} \\
 & BRR-layer & \makebox[41pt][c]{$51.38${\scriptsize$\pm0.30$}} & \makebox[41pt][c]{$71.45${\scriptsize$\pm0.10$}} & \makebox[41pt][c]{$72.01${\scriptsize$\pm0.22$}} & \makebox[41pt][c]{$75.73${\scriptsize$\pm0.18$}} & \makebox[41pt][c]{$63.74${\scriptsize$\pm0.09$}} & \makebox[41pt][c]{$80.65${\scriptsize$\pm0.14$}} & \makebox[41pt][c]{$83.29${\scriptsize$\pm0.05$}} & \makebox[41pt][c]{$83.99${\scriptsize$\pm0.06$}} \\
 & BRR-matrix & \makebox[41pt][c]{$51.50${\scriptsize$\pm0.25$}} & \makebox[41pt][c]{$\mathbf{72.35}${\scriptsize$\pm0.27$}} & \makebox[41pt][c]{$\mathbf{73.16}${\scriptsize$\pm0.14$}} & \makebox[41pt][c]{$\mathbf{76.22}${\scriptsize$\pm0.18$}} & \makebox[41pt][c]{$63.98${\scriptsize$\pm0.10$}} & \makebox[41pt][c]{$\mathbf{80.86}${\scriptsize$\pm0.10$}} & \makebox[41pt][c]{$\mathbf{83.55}${\scriptsize$\pm0.06$}} & \makebox[41pt][c]{$\mathbf{84.31}${\scriptsize$\pm0.11$}} \\
\midrule
 & Channel Merging & \multicolumn{4}{c}{\makebox[41pt][c]{$53.65${\scriptsize$\pm0.14$}}} & \multicolumn{4}{c}{\makebox[41pt][c]{$64.41${\scriptsize$\pm0.05$}}} \\
 & FlexMerge & \makebox[41pt][c]{59.54} & \makebox[41pt][c]{72.11} & \makebox[41pt][c]{69.23} & \makebox[41pt][c]{76.04} & \makebox[41pt][c]{69.88} & \makebox[41pt][c]{81.34} & \makebox[41pt][c]{81.62} & \makebox[41pt][c]{83.60} \\
10 & Shared & \makebox[41pt][c]{$57.40${\scriptsize$\pm0.17$}} & \makebox[41pt][c]{$73.79${\scriptsize$\pm0.27$}} & \makebox[41pt][c]{$73.07${\scriptsize$\pm0.27$}} & \makebox[41pt][c]{$77.81${\scriptsize$\pm0.20$}} & \makebox[41pt][c]{$70.12${\scriptsize$\pm0.26$}} & \makebox[41pt][c]{$82.03${\scriptsize$\pm0.27$}} & \makebox[41pt][c]{$83.85${\scriptsize$\pm0.54$}} & \makebox[41pt][c]{$85.12${\scriptsize$\pm0.51$}} \\
 & BRR-layer & \makebox[41pt][c]{$61.10${\scriptsize$\pm0.43$}} & \makebox[41pt][c]{$77.53${\scriptsize$\pm0.16$}} & \makebox[41pt][c]{$76.31${\scriptsize$\pm0.17$}} & \makebox[41pt][c]{$80.40${\scriptsize$\pm0.11$}} & \makebox[41pt][c]{$74.07${\scriptsize$\pm0.23$}} & \makebox[41pt][c]{$84.77${\scriptsize$\pm0.17$}} & \makebox[41pt][c]{$85.84${\scriptsize$\pm0.08$}} & \makebox[41pt][c]{$86.70${\scriptsize$\pm0.16$}} \\
 & BRR-matrix & \makebox[41pt][c]{$\mathbf{61.87}${\scriptsize$\pm0.15$}} & \makebox[41pt][c]{$\mathbf{78.38}${\scriptsize$\pm0.15$}} & \makebox[41pt][c]{$\mathbf{77.58}${\scriptsize$\pm0.18$}} & \makebox[41pt][c]{$\mathbf{80.83}${\scriptsize$\pm0.07$}} & \makebox[41pt][c]{$\mathbf{74.54}${\scriptsize$\pm0.16$}} & \makebox[41pt][c]{$\mathbf{84.96}${\scriptsize$\pm0.12$}} & \makebox[41pt][c]{$\mathbf{86.13}${\scriptsize$\pm0.10$}} & \makebox[41pt][c]{$\mathbf{86.84}${\scriptsize$\pm0.09$}} \\
\midrule
 & Channel Merging & \multicolumn{4}{c}{\makebox[41pt][c]{$55.29${\scriptsize$\pm0.16$}}} & \multicolumn{4}{c}{\makebox[41pt][c]{$65.67${\scriptsize$\pm0.10$}}} \\
 & FlexMerge & \makebox[41pt][c]{69.82} & \makebox[41pt][c]{76.51} & \makebox[41pt][c]{72.00} & \makebox[41pt][c]{79.18} & \makebox[41pt][c]{79.08} & \makebox[41pt][c]{83.99} & \makebox[41pt][c]{82.45} & \makebox[41pt][c]{85.64} \\
20 & Shared & \makebox[41pt][c]{$67.84${\scriptsize$\pm0.43$}} & \makebox[41pt][c]{$78.02${\scriptsize$\pm0.59$}} & \makebox[41pt][c]{$74.59${\scriptsize$\pm0.22$}} & \makebox[41pt][c]{$80.67${\scriptsize$\pm0.62$}} & \makebox[41pt][c]{$79.36${\scriptsize$\pm0.19$}} & \makebox[41pt][c]{$85.36${\scriptsize$\pm0.35$}} & \makebox[41pt][c]{$85.10${\scriptsize$\pm0.19$}} & \makebox[41pt][c]{$87.03${\scriptsize$\pm0.65$}} \\
 & BRR-layer & \makebox[41pt][c]{$72.57${\scriptsize$\pm0.24$}} & \makebox[41pt][c]{$81.22${\scriptsize$\pm0.19$}} & \makebox[41pt][c]{$78.03${\scriptsize$\pm0.21$}} & \makebox[41pt][c]{$82.62${\scriptsize$\pm0.05$}} & \makebox[41pt][c]{$83.20${\scriptsize$\pm0.20$}} & \makebox[41pt][c]{$87.00${\scriptsize$\pm0.18$}} & \makebox[41pt][c]{$86.61${\scriptsize$\pm0.16$}} & \makebox[41pt][c]{$88.02${\scriptsize$\pm0.12$}} \\
 & BRR-matrix & \makebox[41pt][c]{$\mathbf{74.66}${\scriptsize$\pm0.13$}} & \makebox[41pt][c]{$\mathbf{81.99}${\scriptsize$\pm0.05$}} & \makebox[41pt][c]{$\mathbf{78.99}${\scriptsize$\pm0.23$}} & \makebox[41pt][c]{$\mathbf{83.10}${\scriptsize$\pm0.08$}} & \makebox[41pt][c]{$\mathbf{83.68}${\scriptsize$\pm0.16$}} & \makebox[41pt][c]{$\mathbf{87.18}${\scriptsize$\pm0.11$}} & \makebox[41pt][c]{$\mathbf{86.77}${\scriptsize$\pm0.10$}} & \makebox[41pt][c]{$\mathbf{88.15}${\scriptsize$\pm0.08$}} \\
\bottomrule
\end{tabular*}
\caption{\textbf{CLIP accuracy when merging 100 experts into $K$ modules per unit.} Mean test accuracy (\%) over 100 tasks; randomized methods report mean $\pm$ standard deviation over five seeds, with validation-selected merge scales. FlexMerge and $K=1$ are point estimates. Channel Merging uses its own TA-style merging rule and reports one result per backbone and budget; for bold, it is compared with the TA column. Bold marks the best mean per budget and method column. Paired bootstrap intervals: \cref{app:paired}.}
\label{tab:main}
\end{table*}

\begin{table}[t]
\centering\small
\setlength{\tabcolsep}{5pt}
\begin{tabular}{lcccc}
\toprule
& \multicolumn{2}{c}{ViT-B/32} & \multicolumn{2}{c}{ViT-L/14} \\
\cmidrule(lr){2-3}\cmidrule(lr){4-5}
Method & TA & TSV-M & TA & TSV-M \\
\midrule
Channel Merging & 80.57 & -- & 88.61 & -- \\
FlexMerge & 81.50 & 86.84 & 89.67 & 92.71 \\
Shared & 80.78 & 87.89 & 89.15 & 92.43 \\
BRR-layer & 82.59 & 88.97 & 90.13 & 92.91 \\
BRR-matrix & \textbf{83.12} & \textbf{89.24} & \textbf{90.27} & \textbf{93.00} \\
\bottomrule
\end{tabular}
\caption{\textbf{Public 8-, 14- and 20-task checkpoints.} Mean test accuracy (\%), averaged with equal weight over eight settings: 8 tasks ($K=2,4$), 14 tasks ($K=2,7$) and 20 tasks ($K=2,4,5,10$). Randomized methods use five grouping seeds; Channel Merging uses its own TA-style merging rule and is listed only in the TA columns.}
\label{tab:public-summary}
\end{table}

\begin{table*}[t]
\centering\small
\setlength{\tabcolsep}{1pt}
\begin{tabular*}{\textwidth}{@{\extracolsep{\fill}}cl*{8}{c}@{}}
\toprule
& & \multicolumn{4}{c}{\textbf{InternVL2.5-1B}} & \multicolumn{4}{c}{\textbf{Qwen2-VL-7B}} \\
\cmidrule(lr){3-6}\cmidrule(lr){7-10}
$K$ & Method & TA & TSV-M & Iso-CTS & ACE & TA & TSV-M & Iso-CTS & ACE \\
\midrule
$N$ & Experts & \multicolumn{4}{c}{81.06} & \multicolumn{4}{c}{86.98} \\
\midrule
1 & One model & 66.68 & 74.57 & 75.75 & 76.69 & 79.36 & 85.08 & 84.27 & 86.66 \\
\midrule
 & Channel Merging & \multicolumn{4}{c}{$70.55_{\pm 0.23}$} & \multicolumn{4}{c}{$83.13_{\pm 0.54}$} \\
 & FlexMerge & $70.32$ & $76.85$ & $77.17$ & $78.85$ & $\mathbf{83.94}$ & $86.46$ & $83.99$ & $86.98$ \\
2 & Shared & $71.19_{\pm 0.73}$ & $77.73_{\pm 0.38}$ & $77.29_{\pm 0.20}$ & $79.21_{\pm 0.44}$ & $81.28_{\pm 1.94}$ & $86.25_{\pm 0.19}$ & $84.25_{\pm 0.20}$ & $\mathbf{87.10}_{\pm 0.06}$ \\
 & BRR-layer & $71.94_{\pm 0.08}$ & $78.30_{\pm 0.21}$ & $77.61_{\pm 0.18}$ & $79.59_{\pm 0.10}$ & $83.15_{\pm 0.06}$ & $86.39_{\pm 0.19}$ & $84.44_{\pm 0.08}$ & $87.08_{\pm 0.10}$ \\
 & BRR-matrix & $\mathbf{72.04}_{\pm 0.09}$ & $\mathbf{78.41}_{\pm 0.20}$ & $\mathbf{77.64}_{\pm 0.17}$ & $\mathbf{79.75}_{\pm 0.08}$ & $83.21_{\pm 0.06}$ & $\mathbf{86.48}_{\pm 0.16}$ & $\mathbf{84.48}_{\pm 0.05}$ & $87.09_{\pm 0.06}$ \\
\bottomrule
\end{tabular*}
\caption{\textbf{Vision-language results at $K=2$.} Macro-average test scores (\%) over ten tasks; experts fine-tune all language-model parameters with the visual components frozen. Randomized methods report mean and sample standard deviation over five seeds; FlexMerge and $K=1$ are point estimates. For TA, FlexMerge uses its stronger group-mean scaling (\cref{app:txt-vlm}). Channel Merging uses its own merging rule, with one result per backbone. Bold marks the best mean among FlexMerge, Shared, BRR-layer and BRR-matrix in each method column.}
\label{tab:vlm-fft}
\end{table*}

\section{BRR and Empirical Performance}\label{sec:method}\label{sec:exp}
\subsection{Balanced random regrouping}\label{sec:problem}

\paragraph{Problem setting.}
Let $\theta_1,\ldots,\theta_N$ be $N$ experts fine-tuned from the same pretrained model $\theta_0$, with task vectors $\tau_j=\theta_j-\theta_0$.
We split the network into $U$ units, each a transformer layer or a single weight matrix.
At unit $u$, a grouping $g_u$ divides the experts into $K<N$ groups $G_{u,1},\ldots,G_{u,K}$.
Each group is merged into one module, so every unit keeps $K$ modules and the stored model is as large as $K$ backbones; we call $K$ the budget.

\paragraph{Grouping construction.}
BRR draws the grouping of each unit independently (Fig.~\ref{fig:method}(a)): at unit $u$, it takes a uniformly random permutation of the $N$ experts and splits it into $K$ consecutive groups of equal size $q=N/K$.
This step uses only the expert indices and requires no data, similarity computation or search.
Shared grouping, one of our main baselines, draws one such grouping and reuses it at every unit.
The two schemes therefore have the same group sizes and the same distribution of groupings at each unit; they differ only in whether the units share one grouping.

\paragraph{Merging and inference.}
At unit $u$, the $k$-th module is the pretrained weight plus the merged task vectors of group $G_{u,k}$:
\begin{equation}
\hat{\theta}_{u,k}
=\theta_{0,u}+\alpha\,\mathcal{M}\!\left(\left\{\tau_{j,u}:j\in G_{u,k}\right\}\right).
\label{eq:copy}
\end{equation}
Here, $\theta_{0,u}$ is the pretrained weight at unit $u$ and $\tau_{j,u}$ is the task vector of expert $j$ at that unit; $\mathcal M$ is an existing merging method that maps the task vectors of a group to one merged update, and $\alpha$ scales this update.
Once the pretrained model, the experts, $\mathcal M$ and $\alpha$ are fixed, the modules at unit $u$ depend only on its grouping $g_u$.
BRR changes only this grouping, that is, which task vectors are merged together, not how they are merged.
At inference, the task identity $t$ is known, and each unit $u$ uses the module $\hat\theta_{u,\pi_u(t)}$, where $\pi_u(t)$ is the index of the group containing expert $t$.

\paragraph{Granularity and deployment.}
BRR-layer draws a grouping at each transformer layer, and BRR-matrix at each weight matrix; with six weight matrices per layer, this gives 12 and 72 units for ViT-B/32 and 24 and 144 units for ViT-L/14.
For shared grouping and both BRR variants, each sample passes through one module per unit, so its backbone computation equals that of a single model; any extra cost comes from routing, which depends on the serving implementation (\cref{sec:exp-eff}).

\subsection{Experimental setup}\label{sec:exp-setup}
\paragraph{Tasks and models.}
We fine-tune CLIP~\cite{radford2021clip} ViT-B/32 and ViT-L/14 separately on each of 100 image-classification tasks spanning natural images and specialized domains (\cref{app:tasks}), which gives 100 task experts per backbone.
For these experts, we use budgets $K\in\{5,10,20\}$.
For vision-language evaluation, we use InternVL2.5-1B~\cite{chen2024internvl25} and Qwen2-VL-7B~\cite{wang2024qwen2vl} with budget $K=2$ on ten tasks: six visual question-answering and four multiple-choice image-classification tasks.
Each vision-language expert is obtained by fine-tuning all language-model parameters while keeping the visual components frozen.

\paragraph{Baselines.}
One of our baselines is shared grouping (\cref{sec:problem}), which reuses one grouping at every unit; comparing it with BRR tests whether changing partners across units helps.
The other two, FlexMerge~\cite{dhasade2026flexmerge} and Channel Merging~\cite{zhang2025channel}, group experts by similarity: FlexMerge clusters experts by the similarity of their task vectors in each attention or MLP sublayer, and Channel Merging clusters the corresponding rows of each weight matrix.
BRR, shared grouping and FlexMerge are each combined with four merging methods: Task Arithmetic (TA)~\cite{ilharco2023ta}, TSV-M~\cite{gargiulo2025tsv}, Iso-CTS~\cite{marczak2025iso} and ACE-Merging~\cite{xu2026ace}.
Channel Merging merges each cluster with its own TA-style rule, so it yields one result per backbone and budget.

\paragraph{Evaluation protocol.}
Every method, including FlexMerge and Channel Merging, uses a single merge scale $\alpha$ for all tasks and units, selected on labeled validation data as in Task Arithmetic~\cite{ilharco2023ta} (\cref{app:txt-alpha}).
For randomized methods, $\alpha$ is selected separately for each of five grouping seeds, and we report the mean and standard deviation over seeds of the average test score across tasks.

\subsection{Results on CLIP and VLMs}\label{sec:exp-main}

\paragraph{Partner reuse and grouping granularity.}
\Cref{tab:main} reports CLIP accuracy for two backbones, three budgets and four merging methods; in all 24 configurations, mean accuracy follows BRR-matrix $>$ BRR-layer $>$ shared grouping.
On ViT-B/32 at $K=10$, BRR-matrix improves on shared grouping by 3.02 (ACE-Merging) to 4.59 (TSV-M) points.
These gains are statistically significant in all 24 configurations for both BRR variants (95\% paired bootstrap intervals; \cref{app:paired}).
Regrouping at a finer granularity helps further: BRR-matrix is above BRR-layer in all 24 configurations, by 0.12 to 2.09 points.
The gains persist at a fixed merge scale: with $\alpha=1$ on ViT-B/32, both BRR variants still outperform shared grouping for every merging method and budget (\cref{app:alpha1}).
The gains also hold for pools of 20 and 50 tasks (\cref{app:scale-budget}).

\paragraph{Comparison with existing grouping methods.}
In \cref{tab:main}, BRR-matrix is more accurate than both similarity-based baselines in nearly every setting.
It outperforms FlexMerge in all 24 configurations, by 2.33 (TA) to 8.35 (Iso-CTS) points on ViT-B/32 at $K=10$.
Against Channel Merging, BRR-matrix with TA is more accurate in four of the six backbone--budget settings and less accurate only at $K=5$.
BRR-layer, which regroups only once per layer, also outperforms FlexMerge in all 24 configurations and matches the record of BRR-matrix against Channel Merging.
Shared random grouping already exceeds FlexMerge in 21 of the 24 configurations, and independent regrouping further improves accuracy.

\paragraph{Public checkpoints.}
To test whether the gains depend on our expert pool, we repeat the comparison on the public 8-, 14- and 20-task CLIP checkpoints of Wang et al.~\cite{wang2024tall}.
Averaged over all eight settings, BRR-matrix is the most accurate method in every column of \cref{tab:public-summary}, 0.6--2.3 points above shared grouping and 0.3--2.4 points above FlexMerge (per-setting results: \cref{app:public-bench}).

\paragraph{Vision-language models.}
\Cref{tab:vlm-fft} shows that the gains of BRR carry over to InternVL2.5-1B, whereas on Qwen2-VL-7B they depend on the merging method.
On InternVL2.5-1B, mean scores follow BRR-matrix $>$ BRR-layer $>$ shared grouping for all four merging methods; BRR-matrix improves on shared grouping by 0.35--0.85 points and exceeds FlexMerge in all four settings.
On Qwen2-VL-7B with TA, BRR-matrix is 1.93 points above shared grouping but below FlexMerge.
With TSV-M, Iso-CTS and ACE-Merging, the four grouping methods score almost the same: FlexMerge, shared grouping, BRR-layer and BRR-matrix differ by at most 0.5 points.
For these methods, a single merged model on Qwen2-VL-7B is already within 0.3--2.7 points of the experts, which may explain why grouping methods differ little here.
At a larger budget ($K=5$) and across task types, the gains remain consistent on InternVL2.5-1B and method-dependent on Qwen2-VL-7B (\cref{app:vlm-more}).

\subsection{Effect of balance, independence and similarity}\label{sec:exp-abl}
BRR combines two design choices, balanced group sizes and independent grouping.
\Cref{tab:factor-summary}(a) crosses them on ViT-B/32 with TSV-M and $K=10$, using unbalanced groups of 25 down to 3 experts: independent grouping improves on shared grouping by about 4.5 points with or without balance, and balance adds about 1.4 points in either case (more results in \cref{app:factor-more}).

FlexMerge differs from BRR in two respects at once: its groups are highly unbalanced (at $K=10$, eight experts are kept alone) and its partners are chosen by similarity rather than at random.
To isolate partner selection, we cluster the experts into ten equal-sized groups by task-vector cosine similarity, either once for the whole network or separately for each weight matrix.
Random groups are more accurate in seven of the eight comparisons of \cref{tab:factor-summary}(b), by 1.5--3.9 points; the exception is TA with a shared grouping.
Across 20 comparisons on both backbones, several merging methods and budgets, similarity-based groups are less accurate in 17 (\cref{app:similarity}).
For groupings that vary across units, \cref{tab:partner-repeat} offers one possible explanation: similar experts tend to be grouped together at every unit, so FlexMerge and per-matrix similarity clustering repeat about 30\% of partner pairs between weight matrices, against 9\% for BRR-matrix.
By the analysis of \cref{sec:theory}, such partner reuse can make the errors added at different units accumulate.

\begin{table}[t]
\centering\small
\setlength{\tabcolsep}{4pt}
\begin{tabular}{lcccc}
\toprule
\multicolumn{5}{l}{\emph{(a) Balance and independence (TSV-M)}}\\
Groups & \multicolumn{2}{c}{Shared} & \multicolumn{2}{c}{Per matrix} \\
\midrule
Unbalanced & \multicolumn{2}{c}{72.38} & \multicolumn{2}{c}{76.90} \\
Balanced & \multicolumn{2}{c}{73.79} & \multicolumn{2}{c}{78.38} \\
\midrule
\multicolumn{5}{l}{\emph{(b) Random versus similarity-based partners (balanced)}}\\
& \multicolumn{2}{c}{Shared} & \multicolumn{2}{c}{Per matrix} \\
\cmidrule(lr){2-3}\cmidrule(lr){4-5}
Merging & Random & Similarity & Random & Similarity \\
\midrule
TA & 57.40 & \textbf{57.73} & \textbf{61.87} & 60.37 \\
TSV-M & \textbf{73.79} & 71.22 & \textbf{78.38} & 75.26 \\
Iso-CTS & \textbf{73.07} & 69.67 & \textbf{77.58} & 73.69 \\
ACE & \textbf{77.81} & 75.37 & \textbf{80.83} & 78.54 \\
\bottomrule
\end{tabular}
\caption{\textbf{Effect of balance, independence and similarity.} Mean test accuracy (\%) over 100 tasks with ViT-B/32, $K{=}10$ and validation-selected $\alpha$. Shared reuses one grouping across all matrices; Per matrix samples random groups independently or clusters experts separately at each matrix. Random groupings are averaged over five seeds; similarity-based groupings are deterministic. Full results: \cref{tab:2x2,tab:similarity}.}
\label{tab:factor-summary}
\end{table}

\begin{table}[t]
\centering\small
\setlength{\tabcolsep}{6pt}
\begin{tabular}{lc}
\toprule
Grouping & Partner-repeat rate \\
\midrule
Shared & 1.00 \\
FlexMerge & 0.30 \\
Similarity, per matrix & 0.27 \\
BRR-layer & 0.15 \\
BRR-matrix & 0.09 \\
\bottomrule
\end{tabular}
\caption{\textbf{Partner repetition across units.} Probability that two experts merged together at one weight matrix are also merged together at another, averaged over all pairs of the 72 weight matrices (ViT-B/32, 100 tasks, $K{=}10$). Under BRR-layer, matrices of the same layer share a grouping.}
\label{tab:partner-repeat}
\end{table}

\section{Mechanism Analysis and Validation}\label{sec:mechanism}
\subsection{How changing partners affects feature error}\label{sec:theory}
\Cref{sec:exp} shows consistent accuracy gains from changing partners on CLIP; to understand them, we analyze how merging at different units changes a task's output features.

\paragraph{Feature error.}
For task $t$, we measure the feature error
\begin{equation}
  D_t=\mathbb{E}_x\big\|z(x)-z^\star_t(x)\big\|^2,
  \label{eq:feature-error}
\end{equation}
where $z(x)$ is the $\ell_2$-normalized output of the CLIP ViT image encoder assembled for task $t$ from the merged modules, $z^\star_t(x)$ is that of the task-$t$ expert, and the expectation is over the validation images of task $t$.
CLIP classifies an image by the inner products of $z(x)$ with fixed text embeddings of the class names, so $D_t$ measures how much merging moves the features that the classifier uses.

\paragraph{Local linear approximation.}
Following the tangent-space analysis of Task Arithmetic~\cite{ortiz2023tangent}, we approximate the feature change from merging with its first-order Taylor expansion around the weights of the task-$t$ expert.
At unit $u$, merging changes the expert's weights by $\Delta_u=\hat{\theta}_{u,\pi_u(t)}-\theta_{t,u}$; for brevity, we omit the task subscript on these unit-level quantities.
To first order, this weight change shifts the final normalized feature by $\phi_u(x)=J_u(x)\Delta_u$, where $J_u(x)$ is the Jacobian of $z(x)$ with respect to unit $u$'s weights.
The total first-order feature change is then $\sum_u\phi_u(x)$.

Replacing $z(x)-z^\star_t(x)$ in \cref{eq:feature-error} by this sum gives the linearized feature error
\begin{equation}
\begin{aligned}
D_t^{\mathrm{lin}}&=\mathbb E_x\Big\|\sum_u\phi_u(x)\Big\|^2\\
&=\sum_u\mathbb E_x\|\phi_u(x)\|^2+2\sum_{u<v}\mathbb E_x\langle\phi_u(x),\phi_v(x)\rangle.
\end{aligned}
\label{eq:decomp}
\end{equation}
The first term adds up the error that each unit would cause if it were merged alone.
The second term adds up how the shifts of each pair of units combine: shifts pointing in similar directions ($\mathbb E_x\langle\phi_u(x),\phi_v(x)\rangle>0$) reinforce each other and increase the error, whereas shifts pointing in opposing directions partly offset each other and reduce it.
This is a local approximation; \cref{sec:exp-theory} checks how accurate it is for the weight changes of actual merging.

\paragraph{Covariance from reusing a grouping.}
For a fixed merging method and merge scale, shared and independent grouping draw the grouping at each unit from the same distribution, so the single-unit terms of \cref{eq:decomp} have the same expectation under both; any difference comes from the cross-unit terms.
The shift at unit $u$ depends on its grouping $g_u$, which we make explicit by writing $\phi_u(g_u,x)$.
We split it into its average over random balanced groupings, $\bar\phi_u(x)=\mathbb E_{g_u}[\phi_u(g_u,x)]$, and the remaining deviation $\tilde\phi_u(g_u,x)=\phi_u(g_u,x)-\bar\phi_u(x)$.
Taking the expectation of a cross-unit term over the groupings then gives
\[
\begin{aligned}
&\mathbb E_{g_u,g_v}\mathbb E_x\langle\phi_u(g_u,x),\phi_v(g_v,x)\rangle\\
&\quad=\mathbb E_x\langle\bar\phi_u(x),\bar\phi_v(x)\rangle+\mathbb E_x\mathbb E_{g_u,g_v}\langle\tilde\phi_u(g_u,x),\tilde\phi_v(g_v,x)\rangle.
\end{aligned}
\]
The first term is the same under both groupings.
Under independent grouping, $g_u$ and $g_v$ are drawn independently, so the second term is zero.
Under shared grouping, both units use the same grouping $g$, so the second term becomes the cross-unit covariance
\[
\Gamma_{uv}=\mathbb E_x\mathbb E_g\left\langle\tilde\phi_u(g,x),\tilde\phi_v(g,x)\right\rangle.
\]

\medskip\noindent\textbf{Proposition 1 (covariance identity).}
Let $D_{t,\mathrm{shared}}^{\mathrm{lin}}$ and $D_{t,\mathrm{indep}}^{\mathrm{lin}}$ be the linearized feature errors under shared and independent grouping.
Suppose that the merged modules at each unit depend only on that unit's grouping, and that both groupings use the same merging method, merge scale and grouping distribution at each unit.
Then the difference between their expected values, with expectations over the groupings, is
\begin{equation}
\mathbb E_g\left[D_{t,\mathrm{shared}}^{\mathrm{lin}}\right]
-\mathbb E_g\left[D_{t,\mathrm{indep}}^{\mathrm{lin}}\right]
=2\sum_{u<v}\Gamma_{uv}.
\label{eq:identity}
\end{equation}
The proof is in \cref{app:theory-details}.

A positive $\Gamma_{uv}$ means that, under shared grouping, the deviations at $u$ and $v$ tend to reinforce each other, and a negative value means that they tend to offset each other.
Independent grouping therefore has lower expected linearized feature error exactly when $\sum_{u<v}\Gamma_{uv}>0$, and \cref{sec:exp-theory} tests this condition.

\begin{table*}[t]
\begin{minipage}[t]{0.475\textwidth}\vspace{0pt}
\centering\small
\setlength{\tabcolsep}{5pt}
\begin{tabular}{llccc}
\toprule
\multicolumn{5}{l}{\emph{(a) Relative error $R_\epsilon$ of the first-order approximation}}\\
Method & Grouping & $\epsilon=0.25$ & $\epsilon=0.5$ & $\epsilon=1$ \\
\midrule
TA & Shared & 0.37 & 0.67 & 1.08 \\
 & BRR-layer & 0.37 & 0.67 & 1.07 \\
 & BRR-matrix & 0.37 & 0.67 & 1.07 \\
TSV-M & Shared & 0.30 & 0.56 & 0.95 \\
 & BRR-layer & 0.28 & 0.52 & 0.91 \\
 & BRR-matrix & 0.28 & 0.52 & 0.90 \\
\midrule
\multicolumn{5}{l}{\emph{(b) Partner alignment $\widehat S$}}\\
\multicolumn{4}{l}{Tasks with $\widehat S>0$} & 100/100 \\
\multicolumn{4}{l}{Adjacent layer pairs with $\widehat H_{uv}>0$} & 1100/1100 \\
\bottomrule
\end{tabular}
\caption{\textbf{Checks of the analysis} on 100 ViT-B/32 tasks ($K=10$). (a)~Median over tasks; $R_\epsilon=1$ means the approximation error is as large as the feature change (interquartile ranges: \cref{tab:approx-residual}). (b)~Number of positive estimates, from ten sampled partners and 32 validation images per task.}
\label{tab:checks}
\end{minipage}\hfill
\begin{minipage}[t]{0.5\textwidth}\vspace{0pt}
\centering\small
\setlength{\tabcolsep}{5pt}\renewcommand{\arraystretch}{1.18}
\begin{tabular}{llccc}
\toprule
Unit & Selection objective & \shortstack{Actual\\error $\downarrow$} & \shortstack{Linear.\\error $\downarrow$} & \shortstack{Accuracy\\(\%) $\uparrow$} \\
\midrule
Layer & Random BRR & 0.528 & 0.688 & $77.53\pm0.16$ \\
 & Single-unit & 0.522 & 0.642 & $77.60\pm0.13$ \\
 & Linearized & 0.518 & \textbf{0.631} & $77.89\pm0.10$ \\
 & Actual & \textbf{0.502} & 0.654 & $\mathbf{77.95}\pm0.19$ \\
\midrule
Matrix & Random BRR & 0.478 & 0.626 & $78.38\pm0.15$ \\
 & Single-unit & 0.464 & 0.577 & $78.66\pm0.07$ \\
 & Linearized & 0.458 & \textbf{0.558} & $78.88\pm0.07$ \\
 & Actual & \textbf{0.445} & 0.591 & $\mathbf{78.97}\pm0.07$ \\
\bottomrule
\end{tabular}
\caption{\textbf{Random BRR versus group selection under three objectives} (TSV-M, 100 ViT-B/32 tasks, $K=10$, $\alpha=1.4$). Held-out feature errors and test accuracies are averaged over five seeds; $\pm$ is the standard deviation of accuracy (errors' standard deviations: \cref{tab:search-full}).}
\label{tab:search-main}
\end{minipage}
\end{table*}

\paragraph{Covariance under Task Arithmetic.}
To analyze the covariance further, we consider the most classic merging method, Task Arithmetic (TA), which adds a scaled sum of the group's task vectors to the pretrained weights:
\[
\hat{\theta}_{u,k}=\theta_{0,u}+\lambda\sum_{j\in G_{u,k}}\tau_{j,u},
\]
where $\lambda>0$ is the scaling coefficient of TA.
Each group containing task $t$'s expert also contains $q-1$ partners drawn from the other $N-1$ experts, whose average task vector at unit $u$ is $\bar\tau_u=\frac{1}{N-1}\sum_{j\ne t}\tau_{j,u}$.
We define partner $j$'s contribution at unit $u$ as $\psi_{j,u}(x)=J_u(x)(\tau_{j,u}-\bar\tau_u)$, which measures, to first order, how much choosing partner $j$ changes the output relative to an average partner.

Under shared grouping, task $t$ keeps the same partners at every unit, so we measure how well each partner's contributions at two units $u$ and $v$ align:
\[
H_{uv}=\mathbb E_{j\ne t}\mathbb E_x\left\langle\psi_{j,u}(x),\psi_{j,v}(x)\right\rangle,
\]
which is positive when these contributions tend to point in similar directions.

\medskip\noindent\textbf{Proposition 2 (partner covariance).}
For TA with uniformly sampled balanced groups of size $q=N/K$ and $N>2$,
\begin{equation}
\Gamma_{uv}=cH_{uv},
\qquad c=\lambda^2\frac{(q-1)(N-q)}{N-2},
\label{eq:ta}
\end{equation}
and $c>0$ whenever $1<K<N$, so $\Gamma_{uv}$ and $H_{uv}$ have the same sign; the proof is in \cref{app:ta}.

With $S=\sum_{u<v}H_{uv}$, Propositions~1 and~2 give a gap of $2cS$ in \cref{eq:identity}, so independent grouping has lower expected linearized feature error when $S>0$.
The condition $S>0$ means that the same partner's contributions at different units align positively in aggregate across partners and unit pairs; individual partners or unit pairs may still contribute negatively.
Proposition~2 does not establish that $S$ is positive; \cref{app:kernel} discusses one possible reason, and \cref{sec:exp-theory} estimates $S$ for the actual experts.

\subsection{Verification of the analysis}\label{sec:exp-theory}

\paragraph{First-order approximation.}
Our analysis relies on a first-order approximation, so we first check how well it predicts the change in the output features caused by merging.
On 100 ViT-B/32 tasks with TA and TSV-M at $K=10$, we scale the weight change from merging at every unit by $\epsilon\in\{0.25,0.5,1\}$, giving weights $\theta_{t,u}+\epsilon\Delta_u$, while all other parameters stay at the expert's values; $\epsilon=1$ applies the full change to the grouped weights.
Let $r_\epsilon(x)$ be the actual change in the output feature of image $x$; we compare it with its first-order prediction $\epsilon\sum_u\phi_u(x)$ through the relative error
\[
R_\epsilon=\sqrt{\frac{\mathbb E_{g,x}\big\|r_\epsilon(x)-\epsilon\sum_u\phi_u(x)\big\|^2}{\mathbb E_{g,x}\big\|r_\epsilon(x)\big\|^2}},
\]
averaged over sampled groupings $g$ and 32 validation images per task.
The median $R_\epsilon$ grows from about 0.3 at $\epsilon=0.25$ to about 1 at $\epsilon=1$ (\cref{tab:checks}(a)).
The approximation is thus closer for small weight changes, but at $\epsilon=1$ its error is of the same order as the feature change itself.

\begin{figure*}[t]
\begin{minipage}[t]{0.48\textwidth}
\centering
\includegraphics[height=4.5cm]{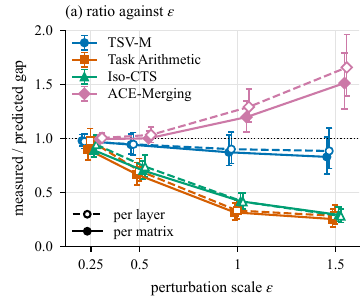}
\caption{\textbf{Ratio of measured to predicted feature-error gaps} (shared minus independent grouping, summed over 100 ViT-B/32 tasks) against the scale $\epsilon$ of the merging-induced weight changes; $\epsilon=1$ applies the full change to the grouped weights, and $\epsilon=1.5$ extrapolates beyond it. A ratio of 1 means agreement. Bars: 95\% task-bootstrap intervals.}
\label{fig:theory-ratio}
\end{minipage}\hfill
\begin{minipage}[t]{0.48\textwidth}
\centering
\includegraphics[height=4.15cm]{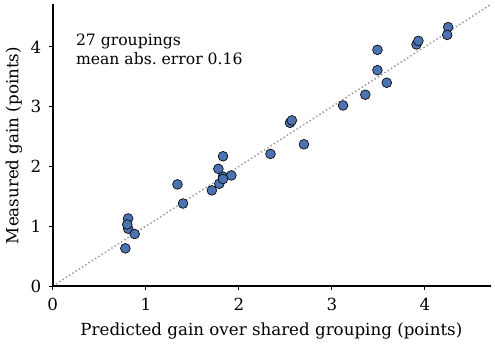}
\caption{\textbf{Predicted versus measured accuracy gain over shared grouping} for 27 groupings on ViT-B/32 (five-seed means). Predictions were fixed in advance and calibrated on the measured gain of BRR over shared grouping, so the shared-grouping and BRR endpoints match by construction; the dotted line marks equality.}
\label{fig:theory}
\end{minipage}
\end{figure*}

\paragraph{Feature-error gap.}
What matters for our analysis, however, is not each feature change but the difference in feature error between shared and independent grouping, which Proposition~1 predicts.
We measure the actual gap as the feature error of \cref{eq:feature-error} under shared grouping minus that under independent grouping.
For each of the four merging methods and each $\epsilon$, we compare it with the first-order prediction $2\epsilon^2\sum_{u<v}\Gamma_{uv}$ of Proposition~1, both summed over the 100 tasks (\cref{fig:theory-ratio}).
In every setting, the actual gap is positive, as predicted: independent grouping lowers the actual feature error, not only its first-order approximation.
Its size also matches closely for small weight changes (ratio 0.89--1.01 at $\epsilon=0.25$) and stays within the same order of magnitude at $\epsilon=1$ (ratio 0.3--1.3).

\paragraph{Partner alignment.}
By Proposition~2, independent grouping lowers the linearized feature error under TA whenever $S=\sum_{u<v}H_{uv}>0$.
Each $H_{uv}$ averages, over partners, the inner product between a partner's contributions at units $u$ and $v$, so $S>0$ means that partners tend to shift the output in similar directions at different units.
Computing $H_{uv}$ requires one Jacobian--vector product per partner and unit, so averaging over all 99 candidate partners is costly.
For each task, we therefore estimate $H_{uv}$ from ten randomly sampled partners on 32 validation images, an unbiased estimate of the average (\cref{app:feature-validation}).
The estimate $\widehat S$ is positive for all 100 tasks, and $\widehat H_{uv}$ is positive for all 11 pairs of adjacent layers in each of the 100 tasks (1100 pairs; \cref{tab:checks}(b)).
These estimates support the positive-alignment condition for our expert pool, and by Propositions~1 and~2, independent grouping lowers the expected linearized feature error under TA, consistent with the positive gap measured above.

\subsection{Applications of the analysis}\label{sec:exp-members}
\paragraph{Predicting the accuracy of partially shared groupings.}
Beyond explaining why BRR outperforms shared grouping, the analysis can predict the accuracy of groupings in between, which share a grouping between some pairs of units but not others.
If units $u$ and $v$ share one grouping draw with probability $p_{uv}$ and draw independently otherwise, the pair adds $2p_{uv}\Gamma_{uv}$ to the expected linearized feature error relative to independent grouping (\cref{sec:exp-dose}); shared grouping has $p_{uv}=1$ for all pairs and BRR has $p_{uv}=0$.
We convert this into accuracy using the measured gain $A$ of BRR over shared grouping: a grouping $a$ is predicted to gain
\[
\widehat A(a)=F(a)\,A,\qquad F(a)=\frac{\sum_{u<v}(1-p_{uv})\,\Gamma_{uv}}{\sum_{u<v}\Gamma_{uv}},
\]
where $F(a)$ is the fraction of the covariance that $a$ removes, so that $F=0$ for shared grouping, $F=1$ for BRR, and groupings in between are interpolated linearly.

We test these predictions on 27 partially shared groupings on ViT-B/32 (\cref{app:predict}).
\Cref{fig:theory} compares each grouping's predicted and measured accuracy gains; points on the diagonal indicate exact agreement.
The measured gains range from 0.6 to 4.3 points, and the predictions are off by 0.16 points on average (results by type: \cref{tab:arrangements}).
These results support using cross-unit covariance to predict the accuracy gains of partially shared groupings.

\paragraph{Searching for better groupings.}
Instead of drawing groupings at random as BRR does, we can choose the groupings that minimize the feature error, estimated either with our first-order analysis or by running the merged model.
At each unit, we draw 64 random permutations to obtain 64 balanced candidate groupings, and select one candidate per unit to minimize an objective summed over all tasks.
We consider three objectives, from coarse to fine: the \emph{single-unit} objective, which scores each unit as if it were the only one merged and so ignores interactions between units; the \emph{linearized} objective, which also includes these interactions; and the \emph{actual} objective, the true feature error.
As there are too many combinations to enumerate, for each objective we first choose candidates unit by unit greedily, then repeatedly revisit each unit and switch it to the candidate that minimizes the objective given the others, until no unit improves.
Comparing the first two objectives shows whether accounting for interactions helps, and comparing the last two shows whether the first-order approximation suffices.
All three use only unlabeled validation images, with TSV-M on 100 ViT-B/32 tasks at $K=10$, five seeds and a fixed merge scale (\cref{app:search-full}).

Accounting for interactions helps (\cref{tab:search-main}): the linearized objective selects more accurate groupings than the single-unit objective in all five seeds, by 0.29 points at layer level and 0.22 at matrix level.
The actual objective adds little further accuracy (0.06--0.09 points), so the first-order approximation largely suffices for this search.
The linearized objective also searches faster, because it computes each candidate's first-order output change once and never runs the merged model during the search, whereas the actual objective runs it for every candidate it considers (\cref{app:search-full}).

Even the best search, however, improves on random BRR by only about half a point, whereas BRR-matrix exceeds shared grouping by 4.59 points (\cref{tab:main}).
Random regrouping thus already obtains most of the gain over shared grouping, while the analysis can still guide a small further improvement.

\section{Conclusion and Future Work}\label{sec:conclusion}

We proposed BRR, which independently draws a balanced random grouping at each layer or weight matrix and merges each group with an existing method; choosing the groups requires no data and only a random permutation of expert indices.
By changing merge partners across layers without increasing the module budget, both variants outperform shared grouping in all 24 CLIP configurations.
BRR-matrix also outperforms FlexMerge in all 24 configurations and with all four merging methods on InternVL2.5-1B.
Our first-order analysis shows that reusing partners introduces correlations between the output-feature shifts caused by merging at different layers.
When these correlations are positive overall, the shifts accumulate; independent regrouping removes these correlations and lowers the expected linearized feature error.
A series of experiments supports this analysis, which further guides a grouping search that can outperform BRR.

\paragraph{Future work.}
An important next step is to account for nonlinear feature changes when predicting and selecting groupings.
Another is to extend BRR to settings where task identities are unavailable, by selecting shared modules from the input.

\section*{LLM Usage Statement}
We used OpenAI Codex and Anthropic Claude to help implement and review code and to polish the language of the manuscript.
\par

{
    \small
    \bibliographystyle{ieeenat_fullname}
    \bibliography{main}
}
\clearpage
\appendix
\makeatletter\newcommand{\inlinetable}{\def\@captype{table}}\setlength{\@fptop}{0pt}\setlength{\@dblfptop}{0pt}\makeatother

\section{Experimental setup and reproducibility}\label{app:setup}
\subsection{Tasks and experts}
\label{app:details}

\label{app:tasks}

For each of the 100 image classification tasks, we separately fine-tuned two CLIP~\cite{radford2021clip} image encoders, ViT-B/32 and ViT-L/14, yielding 100 experts per backbone. Across the 100 tasks, fine-tuning increased mean test accuracy from a zero-shot baseline of 42.05\% to 85.68\% on ViT-B/32 and from 48.11\% to 89.79\% on ViT-L/14. The README in the code repository (\url{https://github.com/kaiyangli-ml/brr-merging}) links to all 200 experts on Hugging Face.

\paragraph{Datasets and splits.}
We obtained 91 tasks from 86 Hugging Face Hub datasets, \texttt{cassava} from the iCassava 2019 release distributed with TFDS, and eight MedMNIST+ tasks from Zenodo. \Cref{tab:tasks} lists the tasks by benchmark suite or application domain and gives the number of classes for each. At the top level of the code repository, \texttt{TASKS\_V3\_100.csv} lists each dataset's source, revision, available license information and split sizes, along with per-task expert and zero-shot test accuracies for both backbones.

For 81 tasks, we held out a stratified 10\% sample of the training split for validation. For another 17 tasks, we used the published validation splits, which hold out entire slides in PCam and entire videos in the KITTI distance task. For the two smallNORB tasks, we kept all images of the same toy in the same split. Within each category, we used all images of toy 9 for validation, toys 4, 6, 7 and 8 for training, and toys 0, 1, 2, 3 and 5 for testing.

For 86 tasks, including both smallNORB tasks, we used the official test split. For another ten tasks, we used a fixed hash-based rule to select 20\% of a published split as the test set. For the remaining four tasks (\texttt{food101}, \texttt{tomato\_leaf}, \texttt{fairface\_age} and \texttt{chess\_pieces}), we used the published validation split for testing and held out a stratified 10\% sample of the training split for validation.

\paragraph{Expert training.}
We fine-tuned only the image encoder and kept the zero-shot classifier built from class-name prompts fixed. We trained each expert for 4,000 steps using Adam with a batch size of 128, a learning rate of $10^{-5}$, a cosine learning rate schedule and no weight decay.

\paragraph{Overlap audit and sensitivity.}
After removing byte-identical images across splits within each task, we compared pixel hashes across all 5.76 million images and found 99 within-task matches between training and test images across three tasks. The only cross-task overlap we detected was between \texttt{tomato\_leaf} and \texttt{plantvillage}, with about 2,500 test images from each task appearing in the other task's training split. In total, five tasks had test images that also appear in a training split, and four of them had more than two each: \texttt{brain\_tumor\_mri}, \texttt{food101}, \texttt{tomato\_leaf} and \texttt{plantvillage}.

On ViT-B/32 with TSV-M and $K=10$, we kept all 100 experts and their merged models fixed and averaged the test accuracy gain of BRR-layer over shared grouping first across all 100 tasks and then across the 95 tasks without detected overlap. Excluding the five affected tasks changed the mean gain from 3.75 to 3.76 percentage points at the validation-selected $\alpha$ and from 2.07 to 2.08 percentage points at $\alpha=1$.

\begin{table*}[h]
\centering\scriptsize
\begin{tabular}{p{0.2\linewidth}p{0.75\linewidth}}
\toprule
Group (count) & Tasks \\
\midrule
FusionBench (TALL20) (20) & sun397 (397), stanford cars (196), resisc45 (45), eurosat (10), svhn (10), gtsrb (43), mnist (10), dtd (47), oxford flowers102 (102), pcam (2), fer2013 (7), oxford pets (37), stl10 (10), cifar100 (100), cifar10 (10), food101 (101), fashion mnist (10), emnist letters (26), kmnist (10), rendered sst2 (2) \\
VTAB / ELEVATER (12) & caltech101 (102), fgvc aircraft (100), country211 (211), clevr count (8), clevr distance (6), dmlab (6), dsprites orientation (40), dsprites location (32), smallnorb azimuth (18), smallnorb elevation (9), diabetic retinopathy (5), kitti distance (4) \\
Fine-grained and food (12) & cub200 (200), stanford dogs (120), birds525 (525), butterflies100 (34), fruits360 (113), pokemon (150), inat wildflowers25 (25), ip102 insects (102), vietnamese food (30), indian food (20), food11 (11), sports100 (100) \\
Objects and scenes (5) & mit indoor67 (67), imagewoof (10), tiny imagenet (200), caltech256 (257), intel scenes (6) \\
DomainNet (6) & domainnet clipart (345), domainnet infograph (345), domainnet painting (345), domainnet quickdraw (345), domainnet real (345), domainnet sketch (345) \\
Remote sensing (4) & whu rs19 (19), rsscn7 (7), patternnet (38), uc merced (21) \\
Agriculture (6) & beans (3), plantvillage (38), cassava (5), cotton leaf (7), tomato leaf (9), fruit ripeness (4) \\
Medical (18) & colorectal histology (8), dermamnist (7), hyper kvasir (23), malaria cells (2), retinamnist (5), breastmnist (2), bloodmnist (8), tissuemnist (8), organamnist (11), organcmnist (11), organsmnist (11), brain tumor mri (4), alzheimer mri (4), chest xray covid (3), dermnet (23), brain hemorrhage (2), lc25000 (5), eeg seizure spectrogram (2) \\
Characters, documents, people, other (17) & usps (10), asl hand gestures (36), bangla vs assamese script (2), rvl cdip (16), galaxy10 decals (10), human action (15), yoga poses (5), fairface age (9), wikiart style (27), wonders of world (12), clothes (7), hagrid gestures (24), deepfake vs real (2), music genre spectrogram (12), chess pieces (13), football players (22), person presence (2) \\
\bottomrule
\end{tabular}
\caption{The 100 tasks of our expert pool, grouped by benchmark suite or application domain. The number after each task name gives its class count.}
\label{tab:tasks}
\end{table*}

\subsection{Grouping baselines and statistics}\label{app:setup-details}

\paragraph{Grouping baselines.}
We compare shared grouping and both BRR variants at the same storage budget, using the same merge operator and the same set of grouping seeds. At each of the 53 module positions in ViT-B/32, we use FlexMerge's~\cite{dhasade2026flexmerge} official greedy complete-linkage procedure to group the corresponding expert modules. For each output channel, Channel Merging clusters the corresponding task-vector rows across experts using spherical $K$-means and merges each cluster using Eq.~(3) of the Channel Merging paper~\cite{zhang2025channel}.

\paragraph{Statistical reporting.}
For each task, we average the accuracy difference between methods over grouping seeds. We count tasks with a positive average difference as improved and, unless otherwise noted, compute 95\% confidence intervals for the mean gain by resampling these task-level differences with replacement (2,000 bootstrap resamples). The intervals for \cref{tab:main} also account for variation across grouping seeds (\cref{app:paired}).

\subsection{Evaluation protocol and scale selection}\label{app:txt-protocol}

\refstepcounter{subsubsection}\label{app:txt-tf32}\paragraph{\thesubsubsection\ Evaluation precision.}
We evaluate all merged models in FP32 with TF32 matrix multiplication disabled.

\refstepcounter{subsubsection}\label{app:txt-alpha}\paragraph{\thesubsubsection\ Scale selection and candidate grids.}
For each CLIP backbone and budget, and for each combination of grouping method, merge operator and applicable grouping seed, we evaluate candidate values of $\alpha$ on a fixed validation sample of up to 1,000 images per task. We select the value with the highest mean validation accuracy across tasks and use it for all tasks and units in that run. The initial grid depends on the merge operator, with shared grouping, BRR and FlexMerge using the same candidates for a given operator.

\begin{center}\small
{\textbf{Initial CLIP merge-scale grids}}\\[3pt]
\begin{tabular}{@{}l@{\hspace{0.8em}}p{0.61\columnwidth}@{}}
\toprule
Method & Initial $\alpha$ grid ($K>1$) \\
\midrule
TSV-M & 0.8, 1.0, 1.2, 1.4, 1.6, 2.0 \\
Task Arithmetic & 1.0, 1.4, 1.6, 2.0, 2.5, 3.0 \\
Iso-CTS & 0.6, 0.8, 1.0, 1.2, 1.4, 1.6, 2.0, 2.5, 3.0 \\
ACE-Merging & 0.6, 0.8, 1.0, 1.2, 1.4, 1.6, 2.0 \\
Channel Merging & 0.25, 0.5, 0.75, 1.0, 1.4, 1.6, 2.0, 2.5, 3.0 \\
\bottomrule
\end{tabular}
\par\smallskip
\begin{minipage}{\columnwidth}\footnotesize
For CLIP Task Arithmetic, $\alpha=1$ gives 1.5 times the group-mean task vector. At $K=1$, the TSV-M, Task Arithmetic and Iso-CTS grids extend down to 0.2.
\end{minipage}
\end{center}

\refstepcounter{subsubsection}\label{app:cm-grid}\paragraph{\thesubsubsection\ Grid-boundary checks.}
TA or Iso-CTS selected the upper bound of $\alpha=3.0$ in 56 runs across 12 configurations; for these we tested three additional values (3.5, 4.0 and 5.0) on the same validation data. Only Iso-CTS on ViT-L/14 at $K=20$ selected a larger scale: BRR-matrix chose 3.5 in all five runs and BRR-layer chose 3.5 in four of five, while shared grouping (all five runs) and FlexMerge retained 3.0. \Cref{tab:main} reports this configuration at the extended selection. TSV-M and ACE-Merging never selected the top of their grids (2.0). In the three configurations where Channel Merging selected the lower bound of 0.25, all 15 runs still selected 0.25 after we extended the grid down to 0.05 in steps of 0.05.

\section{Full results and grouping controls}\label{app:new-controls}
\subsection{Paired accuracy differences}\label{app:paired}
Within each setting of \cref{tab:main}, we computed test accuracy differences between methods on the same task and grouping seed. We resampled both tasks and seeds 4,000 times to estimate a 95\% confidence interval for each mean difference. We summarize these intervals by counting how many lie entirely above zero. For each BRR variant compared with shared grouping, all 24 do. For BRR-matrix over BRR-layer, it lay above zero in 15 settings: 11 of 12 on ViT-B/32 and 4 of 12 on ViT-L/14.

\subsection{Fixed and selected merge scale}\label{app:alpha1}\label{sec:exp-scale}
At a fixed merge scale of $\alpha=1$, both BRR variants achieved higher mean accuracy than shared grouping in all 24 CLIP settings (\cref{tab:alpha1}). With TSV-M, Iso-CTS and ACE-Merging, both variants also outperformed FlexMerge at every budget on both backbones, covering all 18 settings for these three operators. With Task Arithmetic at the same fixed scale, FlexMerge was more accurate than both BRR variants in five of six settings. When each method used its validation-selected scale, both BRR variants outperformed FlexMerge in all six Task Arithmetic settings (\cref{tab:main}).

\begin{table*}[htb]
\centering\footnotesize
\setlength{\tabcolsep}{4pt}
\begin{tabular}{ll cccc c cccc}
\toprule
 & & \multicolumn{4}{c}{ViT-B/32} & & \multicolumn{4}{c}{ViT-L/14} \\
\cmidrule{3-6}\cmidrule{8-11}
$K$ & Method & TA & TSV-M & Iso-CTS & ACE & & TA & TSV-M & Iso-CTS & ACE \\
\midrule
5 & FlexMerge & \textbf{49.74} & 66.90 & 66.13 & 70.36 & & 56.86 & 75.49 & 76.43 & 78.99 \\
 & Shared & 48.73 & 68.32 & 67.27 & 72.65 & & 57.55 & 77.66 & 77.64 & 81.95 \\
 & BRR-layer & 49.05 & 70.54 & 69.08 & 75.73 & & 57.65 & 79.28 & 78.53 & 83.54 \\
 & BRR-matrix & 49.02 & \textbf{71.19} & \textbf{69.49} & \textbf{76.25} & & \textbf{57.69} & \textbf{79.36} & \textbf{78.62} & \textbf{83.74} \\
\midrule
10 & FlexMerge & \textbf{57.67} & 71.58 & 66.29 & 76.04 & & \textbf{65.39} & 79.95 & 76.62 & 83.40 \\
 & Shared & 54.27 & 73.21 & 67.83 & 77.80 & & 64.53 & 81.08 & 77.09 & 85.14 \\
 & BRR-layer & 55.27 & 75.28 & 68.79 & 80.40 & & 65.01 & 82.66 & 77.82 & 86.54 \\
 & BRR-matrix & 55.33 & \textbf{75.82} & \textbf{69.17} & \textbf{80.83} & & 65.16 & \textbf{82.77} & \textbf{77.83} & \textbf{86.66} \\
\midrule
20 & FlexMerge & \textbf{68.14} & 75.63 & 64.56 & 79.18 & & \textbf{75.39} & 82.73 & 73.88 & 85.64 \\
 & Shared & 64.75 & 77.10 & 65.08 & 80.59 & & 73.91 & 84.17 & 75.21 & 87.03 \\
 & BRR-layer & 66.24 & 79.07 & 65.88 & 82.62 & & 74.95 & 85.24 & 75.55 & 88.02 \\
 & BRR-matrix & 66.72 & \textbf{79.47} & \textbf{66.04} & \textbf{83.10} & & 75.10 & \textbf{85.39} & \textbf{75.67} & \textbf{88.15} \\
\bottomrule
\end{tabular}
\caption{Mean test accuracy (\%) over 100 tasks at $\alpha=1$. Random groupings are averaged over five seeds. FlexMerge uses one deterministic grouping. Bold marks the highest accuracy for each backbone, budget and merge operator.}
\label{tab:alpha1}
\end{table*}

\subsection{Balance and independence at $K=10$ on ViT-B/32}
Using TSV-M on 100 ViT-B/32 tasks at $K=10$, we compared shared grouping with independent grouping at each weight matrix under two conditions: equal group sizes and unequal sizes fixed before evaluation (\cref{tab:2x2}). At validation-selected scales, independent grouping improved mean test accuracy by 4.59 points with equal group sizes and by 4.52 points with unequal sizes, improving 95 and 92 of the 100 tasks, respectively. Equal group sizes added 1.41 points under shared grouping and 1.48 points under independent grouping. At fixed $\alpha=1$, independent grouping still added more (2.61 and 2.71 points) than equal group sizes (1.23 and 1.13 points).

\begin{table*}[t]
\centering\normalsize
\setlength{\tabcolsep}{3pt}
\begin{minipage}[t]{0.475\textwidth}
\textbf{(a) Gains at the validation-selected $\alpha$}\par\smallskip
\begin{tabular*}{\linewidth}{@{}l@{\extracolsep{\fill}}rr@{}}
\toprule
Condition & 95\% CI & Tasks improved \\
\midrule
\multicolumn{3}{@{}l}{\textit{Independent minus shared}}\\
Balanced & [3.71, 5.56] & 95/100 \\
Unbalanced & [3.53, 5.54] & 92/100 \\
\addlinespace
\multicolumn{3}{@{}l}{\textit{Balanced minus unbalanced}}\\
Shared & [0.58, 2.18] & 73/100 \\
Independent & [1.14, 1.87] & 91/100 \\
\bottomrule
\end{tabular*}
\end{minipage}\hfill
\begin{minipage}[t]{0.475\textwidth}
\textbf{(b) Accuracy and gains at $K=10$}\par\smallskip
\begin{tabular*}{\linewidth}{@{}l@{\extracolsep{\fill}}rrr@{}}
\toprule
Group sizes & Shared & Independent & Gain \\
\midrule
\multicolumn{4}{@{}l}{\textit{Validation-selected $\alpha$}}\\
Balanced & 73.79 & 78.38 & $+4.59$ \\
Unbalanced & 72.38 & 76.90 & $+4.52$ \\
Balance gain & $+1.41$ & $+1.48$ & -- \\
\addlinespace
\multicolumn{4}{@{}l}{\textit{Fixed $\alpha=1$}}\\
Balanced & 73.21 & 75.82 & $+2.61$ \\
Unbalanced & 71.98 & 74.69 & $+2.71$ \\
Balance gain & $+1.23$ & $+1.13$ & -- \\
\bottomrule
\end{tabular*}
\end{minipage}
\caption{Balance and independent regrouping on ViT-B/32 with TSV-M at $K=10$, 100 tasks, averaged over five grouping seeds. (a) 95\% intervals for gains (points) and counts of improved tasks at validation-selected scales. (b) Mean test accuracy (\%) and gains (points) at selected and fixed scales. Independent grouping draws a partition per matrix. Unequal group sizes are 25, 18, 14, 11, 9, 7, 6, 4, 3 and 3.}
\label{tab:2x2}
\end{table*}

\subsection{Balance and independence across settings}\label{app:factor-controls}\label{app:factor-more}
\Cref{tab:2x2-more} extends this comparison to six additional settings covering other merge operators, budgets and ViT-L/14. In all six settings, both independent grouping and equal group sizes improved mean test accuracy, with a larger gain from independent grouping.

\begin{table}[t]
\centering\small
\setlength{\tabcolsep}{3pt}
\resizebox{\linewidth}{!}{%
\begin{tabular}{llccccc}
\toprule
& & & \multicolumn{2}{c}{Unbalanced sizes} & \multicolumn{2}{c}{Balanced sizes} \\
\cmidrule(lr){4-5}\cmidrule(lr){6-7}
Backbone & Merging & $K$ & shared & independent & shared & independent \\
\midrule
ViT-B/32 & TA & 10 & 56.18 & 61.51 & 57.40 & 61.87 \\
ViT-B/32 & Iso-CTS & 10 & 69.92 & 75.05 & 73.07 & 77.58 \\
ViT-B/32 & ACE-Merging & 10 & 76.07 & 79.80 & 77.81 & 80.83 \\
ViT-B/32 & TSV-M & 5 & 65.57 & 70.16 & 68.32 & 72.35 \\
ViT-B/32 & TSV-M & 20 & 76.89 & 81.18 & 78.02 & 81.99 \\
ViT-L/14 & TSV-M & 10 & 81.01 & 83.88 & 82.03 & 84.96 \\
\bottomrule
\end{tabular}}
\caption{Balance and independent regrouping in six further settings. Mean test accuracy (\%) over 100 tasks at validation-selected scales, averaged over five grouping seeds; independent grouping draws a partition per matrix. At $K=10$, unequal group sizes are as in \cref{tab:2x2}. For $K=5$, adjacent groups are combined (43, 25, 16, 10, 6). For $K=20$, each group is halved as evenly as possible.}
\label{tab:2x2-more}
\end{table}

\subsection{Balanced grouping by task-vector similarity}\label{app:similarity}

With group sizes held equal, does grouping experts by task-vector cosine similarity improve on random grouping? We use negative cosine similarity as the pairwise cost, computed over the whole network for shared grouping or separately within each weight matrix for per-matrix grouping. We embed the cost matrix~\cite{roth2003shift}, run balanced $K$-means from eight initializations, and refine each partition by swapping task pairs. Before evaluating accuracy, we select the partition with the lowest cost.

Clustered and random groupings use the same validation grid. Clustering gives lower mean accuracy in 17 of 20 comparisons (\cref{tab:similarity,tab:similarity-budgets,tab:similarity-l14}). Its three gains all occur with TA and range from 0.28 to 0.61 percentage points. On ViT-B/32 at $K=10$, expert pairs grouped together at one matrix remain together at another in 27\% of cases under per-matrix clustering, compared with 9\% under BRR-matrix (\cref{tab:partner-repeat}). \Cref{app:extensions} gives the expected repeat rate under random grouping.

\begin{table}[htb]
\centering\footnotesize
\setlength{\tabcolsep}{3.5pt}
\begin{tabular}{llcccc}
\toprule
Scope & Members & TA & TSV-M & Iso-CTS & ACE \\
\midrule
Shared & Random & 57.40 & \textbf{73.79} & \textbf{73.07} & \textbf{77.81} \\
 & Similarity & \textbf{57.73} & 71.22 & 69.67 & 75.37 \\
\midrule
Per matrix & Random & \textbf{61.87} & \textbf{78.38} & \textbf{77.58} & \textbf{80.83} \\
 & Similarity & 60.37 & 75.26 & 73.69 & 78.54 \\
\bottomrule
\end{tabular}
\caption{\textbf{Random versus similarity-clustered balanced groups.} Mean test accuracy (\%) on ViT-B/32, 100 tasks, $K=10$, validation-selected $\alpha$. Random groupings are means over five seeds; similarity groupings are deterministic (one run). Bold: the higher of random and similarity grouping for the same grouping scheme and operator.}
\label{tab:similarity}
\end{table}

\begin{table}[htb]
\centering\footnotesize
\setlength{\tabcolsep}{4pt}
\begin{tabular}{lllcc}
\toprule
Method & $K$ & Members & Shared & Per matrix \\
\midrule
TSV-M & 5 & Random & \textbf{68.32} & \textbf{72.35} \\
 & & Similarity & 67.08 & 70.27 \\
 & 20 & Random & \textbf{78.02} & \textbf{81.99} \\
 & & Similarity & 75.00 & 78.53 \\
\midrule
TA & 5 & Random & 50.31 & 51.50 \\
 & & Similarity & \textbf{50.92} & \textbf{51.78} \\
 & 20 & Random & \textbf{67.84} & \textbf{74.66} \\
 & & Similarity & 64.87 & 68.91 \\
\bottomrule
\end{tabular}
\caption{Similarity clustering on ViT-B/32 at $K=5$ and $20$. Values are mean accuracy (\%) on 100 tasks at validation-selected scales. Random groupings use five seeds; similarity clustering uses one deterministic run. Bold: the higher of random and similarity grouping for the same grouping scheme and operator.}
\label{tab:similarity-budgets}
\par\smallskip
\begin{tabular}{llcc}
\toprule
\multicolumn{4}{c}{ViT-L/14, $K=10$}\\
Scope & Members & TA & TSV-M \\
\midrule
Shared & Random & \textbf{70.12} & \textbf{82.03} \\
 & Similarity & 68.57 & 80.07 \\
\midrule
Per matrix & Random & \textbf{74.54} & \textbf{84.96} \\
 & Similarity & 70.77 & 82.86 \\
\bottomrule
\end{tabular}
\caption{Mean test accuracy (\%) for random and similarity-clustered balanced groups on ViT-L/14 at $K=10$, using TA and TSV-M at validation-selected scales. Random groupings use five seeds; similarity clustering uses one deterministic run. Bold: the higher of random and similarity grouping for the same grouping scheme and operator.}
\label{tab:similarity-l14}
\end{table}

\subsection{BRR-matrix gains across merge operators}\label{app:eqbudget-details}
On 100 ViT-B/32 tasks at $K=10$, BRR-matrix improves on shared grouping with all four grouping-only operators (\cref{tab:ops}). At validation-selected scales, the gains range from 3.02 to 4.59 points, every 95\% interval lies above zero, and 93 to 99 of the 100 tasks improve. At $\alpha=1$, the gains are smaller for TA, TSV-M and Iso-CTS (1.06 to 2.61 points) and essentially unchanged for ACE-Merging (3.03 versus 3.02 points). With the mask-based operators, which keep task-specific masks, the gains are 0.72 points with EMR-Merging and 0.20 points with TALL masks, both with intervals above zero (\cref{tab:ops}).

\begin{table}[t]
\centering\small
\setlength{\tabcolsep}{3pt}
\resizebox{\linewidth}{!}{%
\begin{tabular}{lccccc}
\toprule
& \multicolumn{3}{c}{Selected $\alpha$} & & $\alpha{=}1$ \\
\cmidrule(lr){2-4}\cmidrule(lr){6-6}
Merge operator & Shared & BRR-layer & BRR-matrix & gain [95\% CI], tasks improved & gain \\
\midrule
Task Arithmetic & 57.40 & 61.10 & 61.87 & +4.47 [3.61, 5.30], 99/100 & +1.06 \\
TSV-M & 73.79 & 77.53 & 78.38 & +4.59 [3.71, 5.56], 95/100 & +2.61 \\
Iso-CTS & 73.07 & 76.31 & 77.58 & +4.51 [3.62, 5.51], 93/100 & +1.34 \\
ACE-Merging & 77.81 & 80.40 & 80.83 & +3.02 [2.45, 3.68], 94/100 & +3.03 \\
\midrule
EMR-Merging & 81.87 & 82.47 & 82.58 & +0.72 [0.42, 1.00], 77/100 & +0.61 \\
TALL masks & 84.49 & 84.68 & 84.70 & +0.20 [0.07, 0.32], 70/100 & +0.17 \\
\bottomrule
\end{tabular}}
\caption{Mean test accuracy (\%) over 100 ViT-B/32 tasks at $K=10$, averaged over five seeds. Gains compare BRR-matrix with shared grouping (percentage points, computed before rounding). EMR and TALL store additional task-specific masks (\cref{tab:txt-storage}). TA accuracy is lower at $\alpha=1$ for all three groupings (\cref{tab:alpha1,tab:main}).}
\label{tab:ops}
\end{table}

\subsection{Backbone, task counts and budgets}\label{app:scale-budget}\label{sec:exp-second}

\begin{table*}[t]
\centering\small
\setlength{\tabcolsep}{3pt}
\begin{tabular}{llrrcccc}
\toprule
Backbone & Operator & $N$ & $K$ & shared grouping & BRR-matrix & gain [95\% CI] & tasks improved \\
\midrule
ViT-L/14 & TA & 100 & 10 & 70.12 & 74.54 & +4.42 [3.47, 5.36] & 95/100 \\
ViT-L/14 & TSV-M & 100 & 10 & 82.03 & 84.96 & +2.93 & -- \\
ViT-L/14 & Iso-CTS & 100 & 10 & 83.85 & 86.13 & +2.28 [1.58, 3.14] & 88/100 \\
ViT-L/14 & ACE-Merging & 100 & 10 & 85.12 & 86.84 & +1.72 [1.00, 2.57] & 83/100 \\
\midrule
ViT-B/32 & TSV-M & 100 & 2 & 59.71 & 61.71 & +2.00 [1.54, 2.46] & 88/100 \\
ViT-B/32 & TSV-M & 100 & 5 & 68.32 & 72.35 & +4.03 [3.01, 5.05] & 87/100 \\
ViT-B/32 & TSV-M & 100 & 10 & 73.79 & 78.38 & +4.59 [3.71, 5.56] & 95/100 \\
ViT-B/32 & TSV-M & 100 & 20 & 78.02 & 81.99 & +3.97 [3.01, 4.97] & 97/100 \\
ViT-B/32 & TSV-M & 50 & 5 & 68.68 & 73.67 & +4.99 [2.71, 7.87] & 39/50 \\
ViT-B/32 & TSV-M & 50 & 10 & 74.20 & 78.55 & +4.35 [3.03, 5.78] & 45/50 \\
ViT-B/32 & TSV-M & 20 & 2 & 80.23 & 82.36 & +2.12 [0.75, 3.77] & 16/20 \\
ViT-B/32 & TSV-M & 20 & 4 & 84.34 & 86.66 & +2.32 [1.43, 3.31] & 18/20 \\
ViT-B/32 & TSV-M & 20 & 5 & 85.66 & 87.26 & +1.61 [1.01, 2.32] & 18/20 \\
\midrule
ViT-B/32 & Iso-CTS & 100 & 2 & 60.28 & 62.67 & +2.39 [1.82, 2.95] & 90/100 \\
ViT-B/32 & Iso-CTS & 100 & 5 & 68.39 & 73.16 & +4.78 [3.79, 5.76] & 96/100 \\
ViT-B/32 & Iso-CTS & 100 & 20 & 74.59 & 78.99 & +4.40 [3.44, 5.39] & 92/100 \\
ViT-B/32 & Iso-CTS & 50 & 5 & 67.98 & 72.56 & +4.58 [3.46, 5.79] & 47/50 \\
ViT-B/32 & Iso-CTS & 50 & 10 & 70.55 & 75.00 & +4.45 [2.80, 6.26] & 40/50 \\
ViT-B/32 & Iso-CTS & 20 & 2 & 78.56 & 81.49 & +2.93 [0.88, 5.50] & 14/20 \\
ViT-B/32 & Iso-CTS & 20 & 4 & 81.54 & 84.80 & +3.27 [1.69, 5.27] & 18/20 \\
ViT-B/32 & Iso-CTS & 20 & 5 & 81.97 & 84.64 & +2.67 [1.32, 4.53] & 16/20 \\
\midrule
ViT-B/32 & ACE-Merging & 100 & 2 & 63.15 & 65.25 & +2.10 [1.24, 2.97] & 67/100 \\
ViT-B/32 & ACE-Merging & 100 & 5 & 72.65 & 76.22 & +3.57 [2.93, 4.28] & 95/100 \\
ViT-B/32 & ACE-Merging & 100 & 20 & 80.67 & 83.10 & +2.43 [1.64, 3.43] & 84/100 \\
ViT-B/32 & ACE-Merging & 50 & 5 & 72.91 & 77.03 & +4.11 [2.15, 6.26] & 45/50 \\
ViT-B/32 & ACE-Merging & 50 & 10 & 77.83 & 80.16 & +2.32 [1.20, 3.67] & 42/50 \\
ViT-B/32 & ACE-Merging & 20 & 2 & 83.55 & 85.78 & +2.22 [1.28, 3.46] & 20/20 \\
ViT-B/32 & ACE-Merging & 20 & 4 & 86.65 & 88.47 & +1.82 [0.98, 2.86] & 19/20 \\
ViT-B/32 & ACE-Merging & 20 & 5 & 87.46 & 88.75 & +1.29 [0.85, 1.82] & 19/20 \\
\bottomrule
\end{tabular}
\caption{BRR-matrix versus shared grouping at validation-selected scales. Accuracy is the mean test accuracy (\%) over $N$ tasks. Rows with $N=100$ and $K\in\{5,10,20\}$ use five grouping seeds; all other rows use three. Smaller pools use the first $N$ tasks in a fixed order. For TSV-M on ViT-L/14, only the mean gain is available.}
\label{tab:nk}
\end{table*}

\paragraph{Task counts and storage budgets.}
The number of tasks and the storage budget together determine how many partners each expert has at a unit. Across nine combinations of task count and budget on ViT-B/32, BRR-matrix gains 1.61--4.99 points over shared grouping with TSV-M (\cref{tab:nk}). The 95\% interval lies above zero in every setting, and 78--97\% of tasks improve within each setting. Across the 16 ViT-B/32 rows shown for Iso-CTS and ACE-Merging, gains range from 1.29 to 4.78 points, with all intervals above zero.

\paragraph{Larger budgets.}\label{app:largeK}
At $K=25$ on ViT-B/32, BRR-matrix gains 3.20 points over shared grouping with TSV-M and 5.85 with TA. At $K=50$, where each expert has just one partner per unit, the gains are 1.79 with TSV-M and 2.60 with TA, the latter with every TA run at the lower grid boundary $\alpha=1$ (\cref{tab:largeK}).

\begin{table}[htb]
\centering\footnotesize
\setlength{\tabcolsep}{3.5pt}
\begin{tabular}{llccc}
\toprule
$K$ & Method & Shared & BRR-layer & BRR-matrix \\
\midrule
25 & TSV-M & $79.60\pm0.13$ & $82.15\pm0.11$ & $\mathbf{82.80}\pm0.16$ \\
 & TA & $71.55\pm0.13$ & $76.28\pm0.41$ & $\mathbf{77.40}\pm0.11$ \\
\midrule
50 & TSV-M & $82.73\pm0.16$ & $84.34\pm0.12$ & $\mathbf{84.52}\pm0.09$ \\
 & TA$^{\dagger}$ & $80.52\pm0.29$ & $82.71\pm0.09$ & $\mathbf{83.12}\pm0.15$ \\
\bottomrule
\end{tabular}
\caption{\textbf{Larger budgets on ViT-B/32, $N=100$.} Mean test accuracy (\%) and sample standard deviation over three grouping seeds, validation-selected $\alpha$. $\dagger$: at $K=50$, every TA run selected the lower boundary of the grid ($\alpha=1$).}
\label{tab:largeK}
\end{table}

\subsection{Regrouping publicly released CLIP experts}
\label{app:public-bench}

To test whether regrouping also helps with experts trained outside our pipeline, we evaluated TA and TSV-M on the public checkpoints released by Wang et al.~\cite{wang2024tall} for the 8-, 14- and 20-task CLIP suites. Our 100-task pool includes the TALL20 tasks in FusionBench~\cite{tang2024fusionbench}, for which we trained our own experts. We selected merge scales on the public validation splits. The initial grids started at 1.0 for TA and 0.8 for TSV-M. If validation selected the lower bound, we also tested $\{0.4,0.6,0.8\}$ for TA or $\{0.4,0.6\}$ for TSV-M. Whether the released experts saw any of the validation images during training is unknown.

\begin{table*}[tp]
\centering
\footnotesize
\setlength{\tabcolsep}{6pt}\renewcommand{\arraystretch}{0.93}
\caption{Mean test accuracy (\%) on the public 8-, 14- and 20-task suites~\cite{wang2024tall}. Randomized methods report mean $\pm$ standard deviation over five grouping seeds. FlexMerge is deterministic. Channel Merging uses its own merge rule and appears in the TA columns. Bold marks the highest mean among grouping methods for each backbone, operator and budget.}
\label{tab:public-bench}\label{tab:public-bench-l14}\label{tab:public-bench-b32}
\begin{tabular}{@{}ccl cc cc@{}}
\toprule
& & & \multicolumn{2}{c}{ViT-B/32} & \multicolumn{2}{c}{ViT-L/14} \\
\cmidrule(lr){4-5}\cmidrule(lr){6-7}
$N$ & $K$ & Method & TA & TSV-M & TA & TSV-M \\
\midrule
8 & -- & Experts & \multicolumn{2}{c}{92.93} & \multicolumn{2}{c}{95.87} \\
 & -- & Zero-shot & \multicolumn{2}{c}{48.02} & \multicolumn{2}{c}{65.32} \\
 & 1 & One model & 70.64 & 85.62 & 84.84 & 92.96 \\
\cmidrule(l){2-7}
 & 2 & Channel Merging & \ms{84.25}{0.15} & -- & \ms{92.09}{0.08} & -- \\
 &  & FlexMerge & 83.28 & 89.48 & 91.65 & 94.42 \\
 &  & Shared & \ms{81.54}{0.35} & \ms{89.33}{0.30} & \ms{91.86}{0.26} & \ms{94.80}{0.13} \\
 &  & BRR-layer & \ms{83.75}{0.43} & \ms{90.59}{0.08} & \ms{92.86}{0.15} & \ms{95.23}{0.05} \\
 &  & BRR-matrix & \ms{\textbf{84.44}}{0.24} & \ms{\textbf{90.91}}{0.09} & \ms{\textbf{93.05}}{0.12} & \ms{\textbf{95.31}}{0.03} \\
\cmidrule(l){2-7}
 & 4 & Channel Merging & \ms{90.74}{0.09} & -- & \ms{95.00}{0.03} & -- \\
 &  & FlexMerge & 89.24 & 91.31 & 94.23 & 95.13 \\
 &  & Shared & \ms{90.17}{0.17} & \ms{91.94}{0.16} & \ms{95.25}{0.09} & \ms{95.60}{0.05} \\
 &  & BRR-layer & \ms{91.44}{0.10} & \ms{92.51}{0.07} & \ms{95.56}{0.03} & \ms{\textbf{95.83}}{0.02} \\
 &  & BRR-matrix & \ms{\textbf{91.71}}{0.16} & \ms{\textbf{92.60}}{0.10} & \ms{\textbf{95.62}}{0.03} & \ms{95.82}{0.03} \\
\midrule
14 & -- & Experts & \multicolumn{2}{c}{90.94} & \multicolumn{2}{c}{94.33} \\
 & -- & Zero-shot & \multicolumn{2}{c}{57.06} & \multicolumn{2}{c}{68.55} \\
 & 1 & One model & 65.30 & 79.91 & 79.59 & 89.20 \\
\cmidrule(l){2-7}
 & 2 & Channel Merging & \ms{74.81}{0.12} & -- & \ms{85.33}{0.11} & -- \\
 &  & FlexMerge & \textbf{76.28} & 83.36 & \textbf{86.25} & 91.30 \\
 &  & Shared & \ms{73.01}{0.28} & \ms{84.68}{0.44} & \ms{84.74}{0.18} & \ms{91.19}{0.20} \\
 &  & BRR-layer & \ms{74.60}{0.43} & \ms{85.80}{0.13} & \ms{86.10}{0.16} & \ms{92.17}{0.06} \\
 &  & BRR-matrix & \ms{75.04}{0.20} & \ms{\textbf{86.27}}{0.14} & \ms{86.14}{0.14} & \ms{\textbf{92.21}}{0.07} \\
\cmidrule(l){2-7}
 & 7 & Channel Merging & \ms{86.52}{0.20} & -- & \ms{90.96}{0.11} & -- \\
 &  & FlexMerge & 87.78 & 89.25 & 92.74 & 93.47 \\
 &  & Shared & \ms{87.79}{0.36} & \ms{89.57}{0.19} & \ms{93.08}{0.14} & \ms{93.66}{0.05} \\
 &  & BRR-layer & \ms{89.29}{0.22} & \ms{90.17}{0.20} & \ms{93.65}{0.06} & \ms{93.98}{0.05} \\
 &  & BRR-matrix & \ms{\textbf{89.61}}{0.12} & \ms{\textbf{90.37}}{0.12} & \ms{\textbf{93.76}}{0.04} & \ms{\textbf{94.09}}{0.04} \\
\midrule
20 & -- & Experts & \multicolumn{2}{c}{91.41} & \multicolumn{2}{c}{94.76} \\
 & -- & Zero-shot & \multicolumn{2}{c}{56.00} & \multicolumn{2}{c}{65.48} \\
 & 1 & One model & 61.24 & 76.60 & 76.03 & 87.67 \\
\cmidrule(l){2-7}
 & 2 & Channel Merging & \ms{68.10}{0.10} & -- & \ms{\textbf{82.36}}{0.10} & -- \\
 &  & FlexMerge & \textbf{69.63} & 80.60 & 81.93 & 90.12 \\
 &  & Shared & \ms{66.33}{0.57} & \ms{82.50}{1.27} & \ms{80.74}{0.15} & \ms{89.63}{0.53} \\
 &  & BRR-layer & \ms{67.78}{0.25} & \ms{84.38}{0.20} & \ms{81.74}{0.12} & \ms{90.87}{0.31} \\
 &  & BRR-matrix & \ms{68.08}{0.13} & \ms{\textbf{84.60}}{0.21} & \ms{81.78}{0.09} & \ms{\textbf{90.88}}{0.09} \\
\cmidrule(l){2-7}
 & 4 & Channel Merging & \ms{75.43}{0.28} & -- & \ms{85.95}{0.06} & -- \\
 &  & FlexMerge & 77.19 & 84.94 & 88.50 & 91.86 \\
 &  & Shared & \ms{78.06}{1.18} & \ms{86.86}{0.60} & \ms{87.52}{0.83} & \ms{91.42}{1.24} \\
 &  & BRR-layer & \ms{80.14}{0.31} & \ms{88.47}{0.13} & \ms{89.25}{0.15} & \ms{\textbf{91.91}}{0.32} \\
 &  & BRR-matrix & \ms{\textbf{81.19}}{0.32} & \ms{\textbf{88.78}}{0.17} & \ms{\textbf{89.27}}{0.14} & \ms{91.83}{0.13} \\
\cmidrule(l){2-7}
 & 5 & Channel Merging & \ms{78.08}{0.16} & -- & \ms{87.02}{0.07} & -- \\
 &  & FlexMerge & 80.73 & 86.47 & 89.55 & \textbf{92.22} \\
 &  & Shared & \ms{80.76}{0.80} & \ms{88.00}{0.23} & \ms{88.50}{1.08} & \ms{91.19}{1.17} \\
 &  & BRR-layer & \ms{83.80}{0.27} & \ms{89.19}{0.16} & \ms{90.19}{0.15} & \ms{91.55}{0.11} \\
 &  & BRR-matrix & \ms{\textbf{84.60}}{0.16} & \ms{\textbf{89.54}}{0.11} & \ms{\textbf{90.63}}{0.23} & \ms{91.92}{0.19} \\
\cmidrule(l){2-7}
 & 10 & Channel Merging & \ms{86.64}{0.12} & -- & \ms{90.20}{0.09} & -- \\
 &  & FlexMerge & 87.83 & 89.31 & \textbf{92.49} & \textbf{93.15} \\
 &  & Shared & \ms{88.54}{0.30} & \ms{90.23}{0.13} & \ms{91.52}{0.93} & \ms{91.93}{1.07} \\
 &  & BRR-layer & \ms{89.90}{0.15} & \ms{90.67}{0.14} & \ms{91.69}{0.07} & \ms{91.71}{0.07} \\
 &  & BRR-matrix & \ms{\textbf{90.28}}{0.05} & \ms{\textbf{90.88}}{0.09} & \ms{91.94}{0.13} & \ms{91.95}{0.14} \\
\bottomrule
\end{tabular}
\end{table*}

Across eight task-count and budget combinations, two backbones and two merge operators, BRR-matrix achieved higher mean test accuracy than shared grouping in all 32 settings (\cref{tab:public-bench}). BRR-layer did so in 31 settings, with the sole exception of TSV-M on ViT-L/14 at $N=20$ and $K=10$. At least one BRR variant outperformed FlexMerge in 25 of the 32 settings. BRR-matrix with TA outperformed Channel Merging in 14 of 16 comparisons.

\section{Theory and proofs}\label{app:proofs}\label{app:theory-supp}
\subsection{Details for the covariance analysis}\label{app:theory-details}
\paragraph{Proof of Proposition 1 (covariance identity).}
Fix a task $t$, the merging method and the merge scale. With $\bar\phi_u$ and $\tilde\phi_u$ as in \cref{sec:theory}, the first-order feature shift at unit $u$ is $\phi_u(g_u,x)=\bar\phi_u(x)+\tilde\phi_u(g_u,x)$, and the deviation $\tilde\phi_u(g_u,x)$ has mean zero over $g_u$ for each fixed $x$. Both schemes draw $g_u$ from the same distribution, and the update at unit $u$ depends only on $g_u$. Thus the expected single-unit error, $\mathbb E_{g_u}\mathbb E_x\|\phi_u(g_u,x)\|^2$, is the same under both schemes. These terms cancel when we subtract the two expected errors in \cref{eq:decomp}.

Now consider two distinct units $u$ and $v$. For a fixed input $x$, expanding the inner product of their shifts gives four terms. We suppress the arguments for readability.
\[
\begin{aligned}
\langle\phi_u,\phi_v\rangle
={}&\langle\bar\phi_u,\bar\phi_v\rangle+\langle\bar\phi_u,\tilde\phi_v\rangle\\
&+\langle\tilde\phi_u,\bar\phi_v\rangle+\langle\tilde\phi_u,\tilde\phi_v\rangle.
\end{aligned}
\]
The first term is the same under both schemes because each unit has the same mean shift under both. The two mixed terms have expectation zero under either scheme. For example, at fixed $x$, $\bar\phi_u(x)$ is fixed and $\tilde\phi_v(g_v,x)$ has mean zero. Only the last term can differ between the schemes.

Under shared grouping, both units use the same draw $g$. Averaging the last term over groupings and inputs gives the cross-unit covariance defined in \cref{sec:theory},
\[
\mathbb E_x\mathbb E_g
\langle\tilde\phi_u(g,x),\tilde\phi_v(g,x)\rangle=\Gamma_{uv}.
\]
Under independent grouping, the units draw $g_u$ and $g_v$ independently. For each fixed $x$, the expected inner product therefore factors into the inner product of the two means,
\[
\begin{aligned}
&\mathbb E_{g_u,g_v}\langle\tilde\phi_u(g_u,x),\tilde\phi_v(g_v,x)\rangle\\
&\quad=\langle\mathbb E_{g_u}\tilde\phi_u(g_u,x),
\mathbb E_{g_v}\tilde\phi_v(g_v,x)\rangle=0.
\end{aligned}
\]
The shared-minus-independent difference for this pair is therefore $\Gamma_{uv}$. In the squared norm in \cref{eq:decomp}, each unordered pair contributes twice, once in each order. Summing over pairs gives \cref{eq:identity}:
\[
\mathbb E_g[D_{t,\mathrm{shared}}^{\mathrm{lin}}]
-\mathbb E_g[D_{t,\mathrm{indep}}^{\mathrm{lin}}]
=2\sum_{u<v}\Gamma_{uv}.\quad\square
\]

\paragraph{From linearized to true feature error.}
Fix the merge scale $\alpha$ and multiply each weight change from the expert by $\epsilon$. The linearized feature error then scales as $\epsilon^2$. If the output is smooth in $\epsilon$ near the expert ($\epsilon=0$), the true feature-error gap satisfies
\begin{equation}
  \mathbb{E}\big[D_{t,\mathrm{shared}}\big]-\mathbb{E}\big[D_{t,\mathrm{indep}}\big]=2\epsilon^2\sum_{u<v}\Gamma_{uv}+O(\epsilon^3).
  \label{eq:eps}
\end{equation}
Here $\Gamma_{uv}$ uses the full weight displacements ($\epsilon=1$) and the Jacobians evaluated at the expert. The multiplier $\epsilon$ scales the displacement from the expert, whereas $\alpha$ scales the merged update.

\paragraph{Layer and matrix grouping.}
Pairs of matrices in different layers draw independent groupings under both BRR-layer and BRR-matrix, whereas pairs in the same layer share a grouping under BRR-layer but not under BRR-matrix. Summing the covariances of the same-layer pairs gives the expected linearized feature error under BRR-layer minus that under BRR-matrix, $2\sum_b\sum_{m<n\in b}\Gamma_{mn}$, where $b$ indexes layers.

\subsection{Proof of Proposition 2 (partner covariance)}
\label{app:ta}

Fix a task $t$. In a uniformly sampled balanced grouping, its expert belongs to a group of size $q=N/K$. The expert itself is always in the group, so only its partners vary across draws. Let $A$ denote the set of $r=q-1$ partners sampled without replacement from the $M=N-1$ other experts. As in \cref{sec:theory}, let $\bar\tau_u$ be the mean task vector of these $M$ candidates at unit $u$, and write $\psi_{j,u}(x)=J_u(x)(\tau_{j,u}-\bar\tau_u)$ for partner $j$'s contribution relative to that mean. Under TA, the centered feature shift is the scaled sum over the selected partners, while the contributions over all candidates sum to zero:
\[
\tilde\phi_u(g,x)=\lambda\sum_{j\in A}\psi_{j,u}(x),
\qquad \sum_{j\ne t}\psi_{j,u}(x)=0.
\]

Under shared grouping, units $u$ and $v$ use the same partner set $A$. Fix $x$ and omit it from the notation. Expanding the inner product gives
\[
\langle\tilde\phi_u,\tilde\phi_v\rangle
=\lambda^2\sum_{j\in A}\sum_{k\in A}\langle\psi_{j,u},\psi_{k,v}\rangle .
\]
A term with $j=k$ appears whenever partner $j$ is selected, with probability $r/M$. A term with $j\ne k$ requires both partners to be selected, with probability $r(r-1)/[M(M-1)]$.

To combine these terms, sum over all $M$ candidate partners. Since their centered contributions sum to zero at each unit,
\[
\begin{aligned}
0&=\Big\langle\sum_j\psi_{j,u},\sum_k\psi_{k,v}\Big\rangle\\
&=\sum_j\langle\psi_{j,u},\psi_{j,v}\rangle+
\sum_{j\ne k}\langle\psi_{j,u},\psi_{k,v}\rangle .
\end{aligned}
\]
Thus the sum over distinct partners is the negative of the sum over identical partners. Averaging the latter over target images gives $M H_{uv}$, since $H_{uv}$ is the mean alignment over the $M$ candidates. Weighting the two sums by their inclusion probabilities and substituting $r=q-1$ and $M=N-1$ gives \cref{eq:ta}:
\[
\begin{aligned}
\Gamma_{uv}
&=\lambda^2\left[\frac rM-\frac{r(r-1)}{M(M-1)}\right]M H_{uv}\\
&=\lambda^2\frac{r(M-r)}{M-1}H_{uv}\\
&=\underbrace{\lambda^2\frac{(q-1)(N-q)}{N-2}}_{c}\,H_{uv}.
\end{aligned}
\]
For $N>2$, $\lambda>0$ and $1<K<N$, the coefficient $c$ is positive. \hfill$\square$

By Proposition~1, the expected linearized feature error under shared grouping minus that under independent grouping is therefore $2cS$, where $S=\sum_{u<v}H_{uv}$. Independent grouping has lower expected linearized error exactly when $S>0$.

\paragraph{Budget dependence.}
With the coefficient of the group mean held fixed at $\beta=q\lambda$ ($\beta=1.5\alpha$ in our CLIP experiments), the budget $K$ enters through the group size $q=N/K$, giving $c=\beta^2(q-1)(N-q)/[q^2(N-2)]$. For $N=100$, this is $0.010\beta^2$, $0.083\beta^2$ and $0.155\beta^2$ at $K=2$, 10 and 20, so at a fixed merge scale the covariance available to remove depends on $K$.

\subsection{Optimal merge scale}
\label{app:scale}
Proposition~1 compares shared and independent grouping at a fixed merge scale. Here we ask whether independent grouping still gives lower expected feature error when each scheme uses its own optimal scale. We take the pretrained model as the reference and assume that features vary linearly with the weight updates over the scales considered, with the same Jacobians at every scale. Let $e$ be the expert's feature change relative to the pretrained model, and let $a$ be the corresponding change for one random grouping at $\alpha=1$. At scale $\alpha$, the merged model's feature change is $\alpha a$.

Both schemes have the same mean response $\bar a=\mathbb E_g[a]$ because they use the same grouping distribution at each unit. Let $V=\mathbb E_g\|a-\bar a\|^2$ measure how much the response varies across groupings. With squared norms and inner products averaged over target images, the expected feature error is
\[
\mathcal E(\alpha)=\mathbb E_g\|\alpha a-e\|^2
=\|\alpha\bar a-e\|^2+\alpha^2V.
\]
The first term is the same for both schemes. Because the Jacobians do not change with the scale, the argument of Proposition~1 applies to the centered response $a-\bar a$ and gives $V_{\mathrm{shared}}-V_{\mathrm{indep}}=2\sum_{u<v}\Gamma_{uv}$, with the covariances evaluated at $\alpha=1$. When their sum is positive, independent grouping has smaller $V$ and hence lower expected error at every positive scale.

Minimizing $\mathcal E(\alpha)$ gives
\[
\alpha^\star=\frac{\langle\bar a,e\rangle}{\|\bar a\|^2+V},
\qquad
\mathcal E(\alpha^\star)=\|e\|^2-\frac{\langle\bar a,e\rangle^2}{\|\bar a\|^2+V}.
\]
If $\langle\bar a,e\rangle>0$ and both optima lie inside the allowed range, the smaller $V$ under independent grouping gives a larger optimal scale. Its error is already lower at the optimum for shared grouping and decreases further at its own optimum. This calculation minimizes expected feature error over a continuous range of scales. In the experiments, we select the scale from a discrete grid using validation accuracy.

\subsection{Layer alignment at the first step}
\label{app:firststep}
Can a shared training signal make updates at different layers shift the output in similar directions? Consider one gradient descent step, using the same network and images to compute the update and measure its effect on the output. Fix the layer Jacobians and let all layers receive the same output gradient $s$, with mean zero and identity covariance. If $J_u$ is the output Jacobian for layer $u$ and $\eta$ is the step size, updating that layer alone changes the output by $\delta_u=-\eta J_uJ_u^\top s$ to first order. For two layers $u$ and $v$, the expected inner product of their output changes is
\[
\mathbb E_s\langle\delta_u,\delta_v\rangle
=\eta^2\lVert J_u^\top J_v\rVert_F^2\ge 0.
\]
Thus a shared gradient gives a nonnegative interaction on average. If the two layers receive independent zero-mean gradients, the expected interaction is zero. For Gaussian $s$, the first step of bias-corrected Adam gives the same nonnegativity in the sign limit, by Sheppard's formula.

In our experiments, each partner expert trains on its own images, and we measure the effect of its updates on the target task's images. On ViT-B/32, we examined 352 combinations of four partners, eight target tasks and eleven adjacent layer pairs. A sufficient condition for positive interaction is that the Frobenius distance between the training and readout kernels is smaller than the Frobenius norm of the readout kernel. On target images, we measured responses at either the pretrained or the target expert weights, using whole blocks or their second MLP matrices. Across these four settings, the fraction meeting the condition ranged from 2.0\% (target expert, MLP matrices) to 6.5\% (pretrained, whole blocks). With the actual training signal, 98.6--100\% of adjacent-block interactions were positive across both readouts and both update rules (gradient descent and Adam). Positive interactions thus occurred frequently even when the sufficient condition did not hold.

\subsection{Partner repetition under random grouping}\label{app:extensions}

Consider two experts that belong to the same group at one unit. Under shared grouping they remain together at every unit, while under independent balanced grouping each of the $N-1$ other experts is equally likely to occupy one of the $N/K-1$ partner positions at another unit. The probability that the pair remains together is therefore
\begin{equation}
  \frac{N/K-1}{N-1}.
  \label{eq:repeat}
\end{equation}
For $N=100$ and $K=10$, this gives $9/99\approx 9.1\%$, consistent with the 9\% repeat rate measured for BRR-matrix (\cref{tab:partner-repeat}).

\subsection{Understanding positive cross-layer partner interactions}
\label{app:mechanisms}

\subsubsection{Shared training signals and conditional alignment}
\label{app:positive-training}
\label{app:kernel}
For TA with $1<K<N$, the sign of the expected linearized feature-error gap depends on the sum of the partner interactions $H_{uv}$ (\cref{app:ta}). In \cref{app:firststep}, a shared output gradient gives a nonnegative expected interaction at the first step when training and readout use the same network and images. Here a partner trains on its own images, and we ask when two layers transmit differences between partners in similar directions on the target images.

Assume gradient flow from $\theta_0$ with fixed Jacobians and a fixed positive diagonal preconditioner $P$. For partner $j$, let $R_j$ be the output gradient integrated over training on its images $\mathcal D_j$. The update to layer $u$ and its response on the target images $\mathcal X_t$ are
\[
\tau_{j,u}=-P_uJ_u(\mathcal D_j)^\top R_j,\qquad
y_{j,u}=J_u(\mathcal X_t)\tau_{j,u}.
\]
Thus $y_{j,u}=-B_u^{(j)}R_j$, where $B_u^{(j)}=J_u(\mathcal X_t)P_uJ_u(\mathcal D_j)^\top$. Layers share the signal $R_j$ but map it to the target output through different $B_u^{(j)}$.

To compare partners, subtract the mean response of the $M$ candidates at each layer. The interaction $H_{uv}$ averages the inner product of the same partner's centered responses at two layers:
\[
\psi_{j,u}=y_{j,u}-\frac1M\sum_k y_{k,u},\qquad
H_{uv}=\frac1M\sum_j\langle\psi_{j,u},\psi_{j,v}\rangle .
\]
Inner products include the average over target images.

\textbf{Proposition 3.} Let $\mathcal B_u=[B_u^{(1)},\ldots,B_u^{(M)}]$ collect the partner maps for layer $u$. Form one vector per partner by placing $R_j$ in block $j$ and zeros elsewhere, let $\rho_j$ be this vector minus its mean over partners, and let $\Sigma=M^{-1}\sum_j\rho_j\rho_j^\top$. Then
\begin{equation}
\begin{aligned}
H_{uv}&=\operatorname{tr}(\mathcal B_u\Sigma\mathcal B_v^\top),\\
\frac{H_{uv}}{\sqrt{H_{uu}H_{vv}}}
&=\cos\angle(\mathcal B_u\Sigma^{1/2},\mathcal B_v\Sigma^{1/2}).
\end{aligned}
\label{eq:kernel}
\end{equation}
When $H_{uu},H_{vv}>0$, the normalized interaction is positive exactly when the two signal-weighted maps form an acute angle. Positive $S=\sum_{u<v}H_{uv}$ requires positive pair contributions to outweigh negative ones, not every pair to be positive.

\paragraph{Proof.}
With the block vectors, $\mathcal B_u$ applies the matching partner map, so $\psi_{j,u}=-\mathcal B_u\rho_j$. Averaging over partners gives
\[
H_{uv}=\frac1M\sum_j\langle\mathcal B_u\rho_j,\mathcal B_v\rho_j\rangle
=\langle\mathcal B_u\Sigma^{1/2},\mathcal B_v\Sigma^{1/2}\rangle_F .
\]
Writing the Frobenius inner product as a trace gives the first identity. Dividing by the two Frobenius norms gives the second.\hfill$\square$

The derivation holds the Jacobians and preconditioner fixed, whereas both change during Adam training; \cref{sec:exp-theory} therefore measures $S$ directly.

\subsubsection{Signal and partner controls}
\label{app:positive-evidence}
\label{app:mech-controls}
\paragraph{Shared versus independent signals.}
We fix the expert maps and compare the real training signal with two replacements: a random signal shared across layers and independent random signals for each layer. On both backbones, the shared random signal preserves most of the adjacent-layer correlation observed with the real signal, while independent signals give correlations near zero (\cref{tab:shared-signal-controls}).

\begin{table}[H]
\centering\small
\setlength{\tabcolsep}{3pt}
\begin{tabular}{@{}lcc@{}}
\toprule
Output signal & ViT-B/32 & ViT-L/14 \\
\midrule
Real training signal & 0.535 & 0.582 \\
Shared random & $0.517\pm0.016$ & $0.557\pm0.010$ \\
Independent random & $-0.003\pm0.022$ & $0.003\pm0.015$ \\
\bottomrule
\end{tabular}
\caption{\textbf{Adjacent-layer correlation by signal.} Means over 12 tasks per backbone. For fixed-norm random signals, $\pm$ is the mean within-task population SD across four seeds.}
\label{tab:shared-signal-controls}
\end{table}

\paragraph{Same partner, different batches.}
Does the same partner produce aligned layer responses when the training batch changes? For each of 12 target tasks, we compute the first gradient step from $\theta_0$ for ten partners, using batches of 128 training images, and measure how the resulting updates to adjacent fc2 layers affect the target expert's output on 32 validation images. Using different training batches for the same partner retains 89\% of the same-batch interaction (\cref{tab:partner-controls}).

\begin{table}[H]
\centering\small
\setlength{\tabcolsep}{3pt}
\begin{tabular}{@{}p{0.40\columnwidth}p{0.54\columnwidth}@{}}
\toprule
Layer-pair setting & Observed local interaction \\
\midrule
Same partner, same batch & Positive in 132/132 task--layer-pair cases (reference). \\
\addlinespace
Different partners, cyclic shifts & Positive in 2\% of shift cases. \\
\addlinespace
Same partner, different batches & Retains 89\% of the reference interaction; positive in 132/132 cases. \\
\bottomrule
\end{tabular}
\caption{\textbf{Partner alignment across batches.} Same-partner counts cover 132 task--layer pairs (12 tasks, 11 adjacent pairs). The 2\% covers 1,188 cyclic shifts (nine per pair). Centering over ten partners makes the mean shifted interaction $-1/9$ of the same-partner value, so different-partner interactions are negative on average, consistent with only 2\% being positive.}
\label{tab:partner-controls}
\end{table}

\section{Checking the analysis and predicting arrangements}\label{app:checks-predict}
\subsection{Feature-error validation}\label{app:feature-validation}

Proposition~1 (\cref{eq:identity}) gives the expected linearized feature-error gap between shared and independent grouping. We first test how accurately the first-order approximation predicts the feature shift of an individual image. We then compare the gap predicted by Proposition~1 with the difference in actual feature error between the two grouping schemes.

\paragraph{Experimental setup.}
With 100 ViT-B/32 experts and a budget of $K=10$, each balanced group contains ten experts. For each target expert, we draw nine partners at random from the other 99 experts, repeating this procedure to form six candidate groups. Both schemes use these same six groups. Under shared grouping, we use one group at all units and average the results over the six choices. Under independent grouping, each unit chooses uniformly from the six groups, and we average the results over 24 independently assembled models.

\paragraph{Feature shifts and error gaps.}
We scale the weight changes at the regrouped units by $\epsilon$, keeping the remaining weights at the target expert's values. At $\epsilon=1$ the regrouped units receive the full weight change. On 32 validation images per task, we compare the first-order prediction with the actual feature shift. For TA and TSV-M, the median relative residual is close to one at $\epsilon=1$: the prediction error is comparable in size to the feature shift itself (\cref{tab:approx-residual}). The feature-error gap compares two squared errors, so it can still be predicted well when this vector residual is large.

For the gap in \cref{eq:identity}, we subtract the measured feature error under independent grouping from that under shared grouping. We sum the measured and predicted gaps separately over the 100 tasks and compare the two sums. At $\epsilon=0.25$, the ratio of the measured sum to the predicted sum is close to one for all four methods (0.89--1.01). At $\epsilon=1$, independent grouping gives lower measured feature error on at least 87\% of tasks for each method, but the ratio varies by method (0.32--1.29).

\begin{table*}[t]
\centering\small
\begin{tabular}{llccc}
\toprule
Method & Grouping & $\epsilon=0.25$ & $\epsilon=0.5$ & $\epsilon=1$ \\
\midrule
TA & Shared & 0.37 [0.27, 0.47] & 0.67 [0.51, 0.80] & 1.08 [0.91, 1.28] \\
TA & BRR-layer & 0.37 [0.26, 0.46] & 0.67 [0.50, 0.80] & 1.07 [0.89, 1.27] \\
TA & BRR-matrix & 0.37 [0.26, 0.46] & 0.67 [0.50, 0.80] & 1.07 [0.88, 1.27] \\
TSV-M & Shared & 0.30 [0.26, 0.38] & 0.56 [0.48, 0.68] & 0.95 [0.85, 1.11] \\
TSV-M & BRR-layer & 0.28 [0.24, 0.36] & 0.52 [0.45, 0.65] & 0.91 [0.79, 1.06] \\
TSV-M & BRR-matrix & 0.28 [0.23, 0.36] & 0.52 [0.44, 0.64] & 0.90 [0.77, 1.05] \\
\bottomrule
\end{tabular}
\caption{\textbf{Relative output residual of the first-order approximation.} Median [interquartile range] over 100 ViT-B/32 tasks. Shared-grouping results are identical at layer and matrix granularity because every unit uses the same candidate group.}
\label{tab:approx-residual}
\end{table*}

\paragraph{Direct check of Proposition~2.}
On 12 prespecified tasks, covariance sums estimated from 65,536 sampled TA groups per task agree with $cH$ computed from all 99 partners within Monte Carlo uncertainty (\cref{tab:prop2-direct}).

\paragraph{Partner alignment.}
For TA, Propositions~1 and~2 give a shared-minus-independent expected linearized feature-error gap of $2cS$, with $c>0$. Independent grouping therefore gives lower error when $S=\sum_{u<v}H_{uv}$ is positive. We estimate $S$ for each task to test whether feature shifts from the same partner tend to reinforce one another across units in our expert pools.

To reduce computation, we sample ten of the 99 candidate partners per task. At each unit, we subtract the ten partners' mean response from each partner's response, then take the inner product of the same partner's centered responses at units $u$ and $v$. We divide the mean inner product by $(9/10)(99/98)$ to correct the finite-sample bias. This positive correction preserves the sign of $S$. On both backbones, all 100 task estimates are positive, supporting the positive-alignment condition for TA. The tasks share candidate partners, so their estimates are dependent.

\begin{table}[t]
\centering\small
\begin{tabular}{lcccc}
\toprule
Units & \multicolumn{2}{c}{Relative error (\%)} & Sum gap & Max. $|z|$ \\
\cmidrule(lr){2-3}
 & Mean & Max. & (\%) & \\
\midrule
Layer & 0.23 & 0.32 & 0.03 & 1.26 \\
Matrix & 0.30 & 0.43 & 0.02 & 1.17 \\
\bottomrule
\end{tabular}
\caption{\textbf{Direct check of $\Gamma=cH$ on 12 tasks.} We estimate $\Gamma$ from 65,536 balanced TA groups per task (64 batches) and $H$ from all 99 partners ($q=10$, $\lambda=0.15$, 32 images). Relative error is the Frobenius norm of the upper-triangular discrepancy divided by that of $cH$. Sum gap is the mean absolute relative difference between covariance sums. The statistic $z$ is the difference between the estimated and analytic covariance sums divided by the Monte Carlo standard error of that difference.}
\label{tab:prop2-direct}
\end{table}

\subsection{Auxiliary parameters in complete deployments}
\label{app:auxiliary}

Proposition~1 (\cref{eq:identity}) shows how reusing a grouping across units affects expected linearized feature error. In the full model, how much of BRR-layer's accuracy gain comes from grouping the auxiliary parameters separately, and how much comes from changing partners across the main layers?

We start with one grouping shared by all parameters. In the first step, the main layers keep this grouping, and we draw one separate grouping for the embeddings, layer norms and biases to share. This is the $p=1$ control in \cref{sec:exp-dose}. In the second step, we keep the same grouping rule for the auxiliary parameters and draw a grouping independently for each main layer, giving BRR-layer.

On ViT-B/32 with TSV-M at $K=10$ and the validation-selected merge scale, mean accuracy over five seeds changes by $-0.03$ points from fully shared grouping to the $p=1$ control (95\% interval $[-0.52,0.43]$), then rises by 3.78 points from that control to BRR-layer ($[2.92,4.76]$). Changing partners across the main layers therefore accounts for most of the gain in this setting.

\subsection{Predicting grouping arrangements}\label{sec:exp-dose}

\subsubsection{Overview of the 27 arrangements}\label{app:predict}

Can the covariance in Proposition~1 predict accuracy when only some units share a grouping? Let $p_{uv}$ be the probability that units $u$ and $v$ use the same draw, with independent draws otherwise. For two arrangements with the same grouping distribution at each unit, Proposition~1 gives
\begin{equation}
  \mathbb{E}\big[D^{\mathrm{lin}}_{\mathrm{A}}\big]-\mathbb{E}\big[D^{\mathrm{lin}}_{\mathrm{B}}\big]=2\sum_{u<v}\big(p^{\mathrm{A}}_{uv}-p^{\mathrm{B}}_{uv}\big)\Gamma_{uv}.
  \label{eq:arrangement}
\end{equation}

To predict accuracy, we first compute the fraction of shared-grouping covariance that an arrangement removes,
\[
 F=\frac{\sum_{u<v}(1-p_{uv})\Gamma_{uv}}{\sum_{u<v}\Gamma_{uv}}.
\]
If accuracy gain is proportional to this reduction, the predicted gain over shared grouping is $AF$, where $A$ is the measured gain of the corresponding BRR endpoint. The uniform rule weights all pairs equally. The adjacent rule weights only neighboring pairs.

On ViT-B/32 ($N=100$, $K=10$), we made 27 comparisons across four families of grouping arrangements with five grouping seeds. For blocks, placements and matrix ties, we used validation-selected scales and each operator's own covariance and endpoint. Covariances come from \cref{sec:exp-theory}: TSV-M uses all 100 tasks, and TA's layer-pair covariance matrix uses the 20 tasks for which every entry was recorded. The auxiliary parameters follow the default BRR grouping (\cref{app:auxiliary}).

Across the 27 comparisons, the covariance rule predicts accuracy gains with a mean absolute error of 0.16 points, below both comparison rules (\cref{tab:arrangements}). For two TSV-M matrix arrangements (\cref{tab:predict-tie}), all ten runs had finished before the predictions were recorded. One result was available to us and was not used. Across the other 25 comparisons, for which predictions preceded access to results, mean absolute error remains 0.16 points to two decimal places.
\begin{table}[htb]
\centering\small
\setlength{\tabcolsep}{4pt}
\setlength{\tabcolsep}{2.5pt}\begin{tabular}{lcccc}
\toprule
Family & Arms & Covariance & Uniform & Adjacent \\
\midrule
Cycling / copy-refresh & 9 & \textbf{0.17} & 0.68 & 0.61 \\
Blocks of layers & 6 & \textbf{0.17} & 0.95 & 0.85 \\
Placement of changes & 6 & \textbf{0.13} & 1.02 & 0.71 \\
Ties within a layer & 6 & \textbf{0.17} & 0.33 & 2.44 \\
\midrule
All & 27 & \textbf{0.16} & 0.74 & 1.09 \\
\bottomrule
\end{tabular}
\caption{\textbf{Mean absolute prediction error} (accuracy points, ViT-B/32), averaged over the arms of each family. Cycling uses TSV-M; the other families use TSV-M and TA. All rules use the same endpoint calibration. Bold: lowest error in each row. See \cref{app:predict} for the timing of the two additional TSV-M matrix predictions.}
\label{tab:arrangements}
\end{table}

\subsubsection{Results by arrangement}\label{app:x3}

\paragraph{Cycling and copy-refresh.}
Six fixed patterns vary how often the 12 layers cycle through groupings: AAAAAABBBBBB, AAABBBAAABBB, AAAABBBBCCCC, ABABABABABAB, ABCABCABCABC and ABCDABCDABCD. Each letter denotes an independent balanced draw. In copy-refresh, each layer after the first copies the preceding grouping with probability $p$ and draws a fresh one otherwise. Layers $b<c$ therefore share a draw with probability $p^{c-b}$.

Predictions used TSV-M layer covariances and the previously measured BRR-layer endpoint; these predictions and both comparison rules were fixed on 2026-10-30 (SHA-256 \texttt{2acfc8fc}\ldots). Measured gains use the matched $p=1$ control in \cref{app:auxiliary}, with paired gains averaged over five seeds and intervals bootstrapped over tasks (2,000 resamples). More frequent copying lowers mean accuracy at both scales, and the covariance rule has less than half the prediction error of either comparison rule (\cref{tab:x3-full,tab:x3-selected}).
\begin{table*}[t]\centering\small
\begin{tabular}{lcccc}\toprule
Arrangement & Measured [95\% CI] & Covariance & Uniform & Adjacent \\ \midrule
6+6 switch & +0.59 [0.35, 0.84] & +0.45 & +1.13 & +0.19 \\
3+3 & +1.04 [0.79, 1.28] & +0.94 & +1.13 & +0.56 \\
segments of 4 & +0.86 [0.50, 1.19] & +0.78 & +1.51 & +0.38 \\
ABAB & +1.42 [1.13, 1.70] & +1.49 & +1.13 & +2.07 \\
ABC cycle & +1.85 [1.48, 2.22] & +1.86 & +1.51 & +2.07 \\
ABCD cycle & +1.98 [1.60, 2.37] & +1.98 & +1.69 & +2.07 \\
copy-refresh $p{=}0.25$ & +1.91 [1.51, 2.30] & +1.72 & +1.96 & +1.55 \\
copy-refresh $p{=}0.5$ & +1.45 [1.10, 1.79] & +1.29 & +1.76 & +1.03 \\
copy-refresh $p{=}0.75$ & +1.17 [0.86, 1.48] & +0.74 & +1.31 & +0.52 \\
\bottomrule\end{tabular}
\caption{Measured and predicted accuracy gains (points) for nine cycling and copy-refresh arrangements at $\alpha=1$. Measured gains use the matched $p=1$ arm. Predictions use the earlier BRR-layer gain over shared grouping ($+2.07$). A separate five-seed run at $p=0$ gives $+2.09$ ($[1.67,2.53]$). Intervals use paired bootstrap resampling over tasks.}
\label{tab:x3-full}
\end{table*}
\begin{table*}[t]\centering\small
\begin{tabular}{lcccc}\toprule
Arrangement & Measured [95\% CI] & Covariance & Uniform & Adjacent \\ \midrule
6+6 switch & +0.96 [0.68, 1.27] & +0.81 & +2.04 & +0.34 \\
3+3 & +1.60 [1.24, 1.96] & +1.71 & +2.04 & +1.02 \\
segments of 4 & +1.38 [0.95, 1.85] & +1.40 & +2.72 & +0.68 \\
ABAB & +2.37 [1.75, 3.10] & +2.70 & +2.04 & +3.75 \\
ABC cycle & +3.20 [2.42, 4.10] & +3.36 & +2.72 & +3.75 \\
ABCD cycle & +3.40 [2.64, 4.27] & +3.59 & +3.06 & +3.75 \\
copy-refresh $p{=}0.25$ & +3.02 [2.27, 3.91] & +3.12 & +3.54 & +2.81 \\
copy-refresh $p{=}0.5$ & +2.21 [1.60, 2.86] & +2.34 & +3.18 & +1.87 \\
copy-refresh $p{=}0.75$ & +1.70 [1.23, 2.22] & +1.34 & +2.36 & +0.94 \\
\bottomrule\end{tabular}
\caption{Measured and predicted accuracy gains (points) for nine cycling and copy-refresh arrangements at validation-selected scale. Measured gains use the matched $p=1$ arm. Predictions use the earlier BRR-layer gain over shared grouping ($+3.75$). Relative to the matched $p=1$ arm, the same BRR-layer run gives $+3.78$ (\cref{app:auxiliary}), and a separate five-seed run at $p=0$ gives $+3.55$ ($[2.75,4.43]$). Intervals use paired bootstrap resampling over tasks.}
\label{tab:x3-selected}
\end{table*}

Using the same nine arrangements and TSV-M fractions $F$, we calibrated predictions to the ACE-Merging and Iso-CTS endpoints before evaluation. The covariance rule has the lowest error for both operators at the selected scale and for ACE-Merging at $\alpha=1$. It ties the uniform rule for Iso-CTS at $\alpha=1$ (\cref{tab:x3-ops}).
\begin{table}[t]
\centering\small
\setlength{\tabcolsep}{3pt}
\begin{tabular*}{\columnwidth}{@{\extracolsep{\fill}}llccc@{}}
\toprule
Operator & Scale & Covariance & Uniform & Adjacent \\
\midrule
ACE-Merging & $\alpha{=}1$ & \textbf{0.16} & 0.40 & 0.52 \\
ACE-Merging & selected & \textbf{0.15} & 0.40 & 0.51 \\
Iso-CTS & $\alpha{=}1$ & 0.16 & 0.16 & 0.28 \\
Iso-CTS & selected & \textbf{0.32} & 0.53 & 0.64 \\
\bottomrule
\end{tabular*}
\caption{\textbf{Mean absolute error of predictions fixed in advance for other operators.} Accuracy points over the nine arrangements, ViT-B/32, 100 tasks, $K{=}10$. Bold: lowest error in a row when it is unique.}
\label{tab:x3-ops}
\end{table}

\paragraph{Blocks of layers.}
Shortening independently grouped blocks from six layers to three and then two increases mean accuracy for both operators, as predicted (\cref{tab:predict-blk}). Three comparisons among the 27 repeat a design with fresh draws: the six-layer TSV-M block repeats the 6+6 switch, and evenly spaced changes repeat the three-layer block for both operators.
\begin{table}[htb]
\centering\footnotesize
\setlength{\tabcolsep}{3pt}
\footnotesize\setlength{\tabcolsep}{2pt}\begin{tabular}{llcccc}
\toprule
Method & Layers/block & Measured & Covariance & Uniform & Adjacent \\
\midrule
TSV-M & 6 & $74.92\pm0.30$ & 74.60 & 75.83 & 74.13 \\
 & 3 & $75.62\pm0.20$ & 75.62 & 76.85 & 74.81 \\
 & 2 & $76.52\pm0.28$ & 76.34 & 77.19 & 75.49 \\
\midrule
TA & 6 & $58.43\pm0.25$ & 58.20 & 59.42 & 57.74 \\
 & 3 & $59.11\pm0.52$ & 59.19 & 60.43 & 58.41 \\
 & 2 & $60.17\pm0.36$ & 59.97 & 60.76 & 59.08 \\
\bottomrule
\end{tabular}
\caption{\textbf{Blocks of consecutive layers sharing a grouping.} Measured mean test accuracy (\%) and sample standard deviation over five seeds, and the accuracies predicted by the covariance rule (fixed before measurement) and by the two comparison rules. Endpoints: shared grouping 73.79 (TSV-M) and 57.40 (TA); BRR-layer 77.53 and 61.10.}
\label{tab:predict-blk}
\end{table}

\paragraph{Placement of three changes.}
Here every arrangement has exactly three changes of grouping along the 12 layers, so the layers form four contiguous segments.
For each operator, we compare the placements with the largest and smallest predicted gains among the 165 possible placements, together with evenly spaced changes. For TSV-M, the predicted-best placement beats the predicted-worst in every seed but has no consistent advantage over even spacing. The uniform rule also captures the best-to-worst gap and is closer for TA; the adjacent rule predicts no difference (\cref{tab:predict-place}).
\begin{table}[htb]
\centering\footnotesize
\setlength{\tabcolsep}{2.5pt}\begin{tabular}{lcccc}
\toprule
Layers per segment & Measured & Covariance & Uniform & Adjacent \\
\midrule
\multicolumn{5}{l}{\textit{TSV-M}} \\
Best (4, 2, 2, 4) & $75.64\pm0.18$ & 75.71 & 76.74 & 74.81 \\
Even (3, 3, 3, 3) & $75.58\pm0.33$ & 75.62 & 76.85 & 74.81 \\
Worst (1, 9, 1, 1) & $74.42\pm0.40$ & 74.57 & 75.49 & 74.81 \\
\midrule
\multicolumn{5}{l}{\textit{TA}} \\
Best (4, 2, 3, 3) & $59.57\pm0.34$ & 59.23 & 60.37 & 58.41 \\
Even (3, 3, 3, 3) & $59.36\pm0.48$ & 59.18 & 60.43 & 58.41 \\
Worst (1, 1, 9, 1) & $58.27\pm0.45$ & 58.28 & 59.08 & 58.41 \\
\bottomrule
\end{tabular}
\caption{\textbf{Three partner changes at different layers} (ViT-B/32, $K=10$). Measured mean test accuracy (\%) and sample standard deviation over five seeds, with the covariance, uniform and adjacent predictions fixed before measurement. Best and worst are the placements with the largest and smallest predicted $F$ for each operator. Endpoints: shared grouping 73.79 (TSV-M) and 57.40 (TA); BRR-layer 77.53 and 61.10.}
\label{tab:predict-place}
\end{table}

\paragraph{Tying matrices within a layer.}
With matrices as units, the six weight matrices of every layer (in forward order: query, key, value and output projections, then the two MLP matrices) are tied into sets that share a grouping; all sets are independent of one another.
We considered all five ways to split the six matrices into two contiguous sets. Using only the $72\times72$ covariance, we selected the split with the largest $F$ (the second MLP matrix alone) and the split with the smallest $F$ (the query projection alone).
Each split isolates one end matrix, so the two splits tie the same number of pairs overall and the same number of adjacent pairs. Each comparison rule therefore predicts the same gain for the two splits.

For both TSV-M and TA, the covariance rule favors isolating the second MLP matrix, and this split wins in every seed. For TSV-M the rule overestimates the difference (\cref{tab:predict-tie}).
\begin{table}[htb]
\centering\footnotesize
\setlength{\tabcolsep}{2.5pt}\begin{tabular}{lcccc}
\toprule
Tied sets & Measured & Covariance & Uniform & Adjacent \\
\midrule
\multicolumn{5}{l}{\textit{TSV-M}} \\
Best (fc2 alone) & $78.12\pm0.12$ & 78.04 & 78.16 & 75.28 \\
Worst (query alone) & $77.74\pm0.12$ & 77.28 & 78.16 & 75.28 \\
Two sets & $77.83\pm0.19$ & 77.70 & 78.23 & 75.28 \\
Three sets & $77.89\pm0.16$ & 77.72 & 78.32 & 76.05 \\
\midrule
\multicolumn{5}{l}{\textit{TA}} \\
Best (fc2 alone) & $61.60\pm0.16$ & 61.64 & 61.66 & 58.85 \\
Worst (query alone) & $61.01\pm0.21$ & 60.89 & 61.66 & 58.85 \\
\bottomrule
\end{tabular}
\caption{\textbf{Tying the matrices of a layer} (ViT-B/32, $K=10$). Mean test accuracy (\%) and sample standard deviation over five seeds, with predictions fixed before the runs (for the TSV-M two- and three-set arrangements, see \cref{app:predict} for timing). Two sets: the four attention matrices and the two MLP matrices; three sets: query and key, value and output, and the two MLP matrices. Endpoints: shared grouping 73.79 (TSV-M) and 57.40 (TA); BRR-matrix 78.38 and 61.87.}
\label{tab:predict-tie}
\end{table}
For the additional two- and three-set TSV-M arrangements in \cref{tab:predict-tie}, both predicted and measured accuracies were close.

\paragraph{Placement of a single switch.}
To test whether the ranking of the cycling arrangements above can be explained by the number of switches alone, we use two groupings and exactly one switch in every arrangement: layers $1,\dots,s$ use grouping A and the remaining layers use grouping B, with $s\in\{2,4,6,8,10\}$.
For each seed, these arrangements reuse the first two cycling groupings, so $s{=}6$ coincides with the 6+6 switch. These five arrangements are not among the 27 arms and are evaluated at $\alpha=1$ only.

At fixed budget, the covariance rule correctly orders the mean gains for eight of ten pairs of switch points, but its task-level success rate is only 61\% (ties excluded, interval $[0.57,0.66]$). A post-hoc rule using only the covariance across the switch has the same rank correlation but larger magnitude errors (\cref{tab:eqb}).
\begin{table}[h]
\centering\small
\setlength{\tabcolsep}{4pt}
\resizebox{\columnwidth}{!}{%
\begin{tabular}{lcc}
\toprule
prediction & MAE, error ratio & MAE, accuracy (pp) \\
\midrule
covariance $F(s)$ (fixed in advance) & \textbf{0.029} & \textbf{0.113} \\
uniform rule (fixed in advance) & 0.268 & 0.504 \\
adjacent rule (fixed in advance) & 0.076 & 0.239 \\
adjacent pair, $\Gamma$-weighted (post hoc) & 0.065 & 0.214 \\
\bottomrule
\end{tabular}}
\caption{\textbf{Where to switch: the covariance prediction has the smallest error.}
Equal-budget arrangements (two groupings, one switch at $s\in\{2,4,6,8,10\}$), TSV-M, ViT-B/32, 100 tasks, $K{=}10$, five seeds, $\alpha{=}1$. Error ratio: feature-error gap normalized by the gap between the shared grouping and BRR-layer. Bold: lowest error in each column.}
\label{tab:eqb}
\end{table}

\paragraph{Rules evaluated after the runs.}
With the same endpoint calibration, a rule using only layer distance has a mean absolute error of 0.23 points across all 27 comparisons, against 0.16 for covariance. For the matrix ties tested here, matrix type and layer distance reproduce the full covariance predictions (\cref{tab:null-rules}).
\begin{table}[htb]
\centering\footnotesize
\setlength{\tabcolsep}{2.5pt}
\resizebox{\columnwidth}{!}{%
\begin{tabular}{lcccccccc}
\toprule
 & & \multicolumn{3}{c}{fixed in advance} & \multicolumn{4}{c}{post hoc} \\
\cmidrule(lr){3-5}\cmidrule(lr){6-9}
Family & $n$ & Cov. & Unif. & Adj. & Avg. & Dist. & Type+dist. & Geom. \\
\midrule
Nine arrangements (TSV-M) & 9 & 0.17 & 0.68 & 0.61 & 0.19 & 0.19 & 0.19 & 0.15 \\
Blocks, TSV-M & 3 & 0.17 & 0.94 & 0.88 & 0.15 & 0.25 & 0.25 & 0.19 \\
Blocks, TA & 3 & 0.17 & 0.97 & 0.83 & 0.23 & 0.22 & 0.22 & 0.17 \\
Placement, TSV-M & 3 & 0.09 & 1.15 & 0.66 & 0.37 & 0.23 & 0.23 & 0.33 \\
Placement, TA & 3 & 0.18 & 0.89 & 0.75 & 0.24 & 0.30 & 0.30 & 0.19 \\
Ties, TSV-M & 2 & 0.27 & 0.23 & 2.65 & 1.21 & 0.32 & 0.27 & -- \\
Ties, TA & 2 & 0.08 & 0.36 & 2.46 & 1.05 & 0.30 & 0.08 & -- \\
Second tie pair, TSV-M & 2 & 0.15 & 0.41 & 2.20 & 0.89 & 0.14 & 0.15 & -- \\
\midrule
All & 27 & 0.16 & 0.74 & 1.09 & 0.41 & 0.23 & 0.21 & 0.19$^\ast$ \\
\bottomrule
\end{tabular}}
\caption{\textbf{Mean absolute error of the covariance rule and of simpler rules} on ViT-B/32 (points; all rules use the same endpoint calibration). Cov., Unif., Adj.: the covariance prediction and the uniform and adjacent rules. See \cref{app:predict} for prediction timing in the second TSV-M tie pair. The remaining columns were computed after all accuracies were known: Avg., the mean of the uniform and adjacent predictions; Dist., $\Gamma_{uv}$ replaced by its mean over unit pairs at the same layer distance; Type+dist., by its mean over pairs with the same matrix types and layer distance (equal to Dist.\ for layer-level arrangements); Geom., $\Gamma_{bc}\propto 0.6^{|b-c|}$ (layer-level arrangements only). $^\ast$: the 21 layer-level arms only.}
\label{tab:null-rules}
\end{table}

\section{Analysis-guided group selection}\label{app:search-full}

\paragraph{Search setup.}
We test whether accounting for interactions between units helps select better groupings. At each unit, either a layer or a weight matrix, all three objectives choose from the same 64 balanced candidate groupings, including the one used by random BRR. Each search uses 16 unlabeled validation images per task. We measure feature error on a separate set of 16 images per task and classification accuracy on the test sets. We use TSV-M on 100 ViT-B/32 tasks at $K=10$, with fresh candidates for each of five seeds. All comparisons use $\alpha=1.4$. Validation selected this scale in every run except one matrix-level search with the actual objective, which selected 1.6 (\cref{tab:search-full}).

Let $\phi_{t,u}(x)$ be the first-order change in the final normalized feature of expert $t$ when unit $u$ is merged. The single-unit objective sums the squared feature shifts, so each unit can choose its grouping independently. The full linearized objective also accounts for how shifts from different units interact (\cref{eq:decomp}):
\[
\begin{aligned}
\mathcal L_{\mathrm{full}}={}&\underbrace{\frac{1}{N}\sum_t\sum_u\mathbb E_x
\|\phi_{t,u}(x)\|^2}_{\mathcal L_{\mathrm{unit}}}\\
&+\frac{2}{N}\sum_t\sum_{u<v}\mathbb E_x
\langle\phi_{t,u}(x),\phi_{t,v}(x)\rangle .
\end{aligned}
\]
The third objective runs the assembled model and measures its actual feature error relative to each task expert (\cref{eq:feature-error}), averaged over tasks. For the full linearized and actual objectives, we initialize the search greedily, then revisit one unit at a time with the other groupings fixed until a full pass changes no grouping.

\begin{table*}[htb]
\centering\small
\setlength{\tabcolsep}{6pt}
\begin{tabular}{llccc}
\toprule
Granularity & Selection & \shortstack{Actual held-out\\feature error} & \shortstack{Linearized held-out\\feature error} & \shortstack{Test accuracy\\(\%)} \\
\midrule
Layer & Random BRR & $0.5283\pm0.0019$ & $0.6875\pm0.0081$ & $77.53\pm0.16$ \\
 & Single-unit objective & $0.5216\pm0.0044$ & $0.6423\pm0.0015$ & $77.60\pm0.13$ \\
 & Linearized objective & $0.5177\pm0.0044$ & $\mathbf{0.6312}\pm0.0034$ & $77.89\pm0.10$ \\
 & Actual objective & $\mathbf{0.5023}\pm0.0026$ & $0.6543\pm0.0116$ & $\mathbf{77.95}\pm0.19$ \\
\midrule
Matrix & Random BRR & $0.4776\pm0.0045$ & $0.6262\pm0.0022$ & $78.38\pm0.15$ \\
 & Single-unit objective & $0.4641\pm0.0023$ & $0.5765\pm0.0031$ & $78.66\pm0.07$ \\
 & Linearized objective & $0.4583\pm0.0012$ & $\mathbf{0.5579}\pm0.0048$ & $78.88\pm0.07$ \\
 & Actual objective & $\mathbf{0.4452}\pm0.0035$ & $0.5912\pm0.0026$ & $\mathbf{78.97}\pm0.07$ \\
\bottomrule
\end{tabular}
\caption{\textbf{Group selection under three objectives.} Held-out feature errors and test accuracy at fixed $\alpha=1.4$, reported as means and sample standard deviations over five seeds. Bold marks the lowest mean error or highest mean accuracy at each granularity. In the one matrix-level run that selected $\alpha=1.6$, test accuracy was 78.89\% at 1.6 and 78.99\% at 1.4.}
\label{tab:search-full}
\end{table*}

\paragraph{Selection results.}
Including cross-unit interactions improves test accuracy over single-unit selection in all five seeds at both granularities (\cref{tab:search-paired}). Selecting groups by actual feature error reduces held-out actual error further in every seed, but adds only 0.06 percentage points to mean test accuracy for layers and 0.09 for matrices. The selected models also have higher mean linearized error than those chosen by the linearized objective (\cref{tab:search-full}). Thus, the two objectives favor different groupings even though their mean test accuracies are close.

\begin{table*}[t]
\centering\small
\setlength{\tabcolsep}{5pt}
\begin{tabular}{llcccc}
\toprule
 & & \multicolumn{2}{c}{Test accuracy (points)} & \multicolumn{2}{c}{Actual held-out feature error} \\
\cmidrule(lr){3-4}\cmidrule(lr){5-6}
Granularity & Comparison & Difference & Higher in & Difference & Lower in \\
\midrule
Layer & Linearized $-$ random & $+0.36\pm0.18$ & 5/5 & $-0.0105\pm0.0042$ & 5/5 \\
 & Single-unit $-$ random & $+0.06\pm0.18$ & 2/5 & $-0.0067\pm0.0033$ & 5/5 \\
 & Linearized $-$ single-unit & $+0.30\pm0.12$ & 5/5 & $-0.0038\pm0.0063$ & 4/5 \\
 & Actual $-$ random & $+0.42\pm0.18$ & 5/5 & $-0.0260\pm0.0027$ & 5/5 \\
 & Actual $-$ linearized & $+0.06\pm0.18$ & 3/5 & $-0.0154\pm0.0034$ & 5/5 \\
\midrule
Matrix & Linearized $-$ random & $+0.50\pm0.16$ & 5/5 & $-0.0193\pm0.0055$ & 5/5 \\
 & Single-unit $-$ random & $+0.28\pm0.20$ & 4/5 & $-0.0135\pm0.0043$ & 5/5 \\
 & Linearized $-$ single-unit & $+0.22\pm0.08$ & 5/5 & $-0.0058\pm0.0024$ & 5/5 \\
 & Actual $-$ random & $+0.59\pm0.18$ & 5/5 & $-0.0323\pm0.0050$ & 5/5 \\
 & Actual $-$ linearized & $+0.09\pm0.03$ & 5/5 & $-0.0130\pm0.0040$ & 5/5 \\
\bottomrule
\end{tabular}
\caption{\textbf{Paired differences between selection objectives.} Differences are computed within each seed at $\alpha=1.4$ and reported as mean $\pm$ sample standard deviation. Counts show how often the first objective gives higher accuracy or lower error. For layer-level accuracy, the mean paired gain of linearized over single-unit selection is 0.30 points. Subtracting the rounded means in \cref{tab:search-main} gives 0.29.}
\label{tab:search-paired}
\end{table*}

\paragraph{Search cost.}
Linearized search computes each candidate's feature shift once and reuses it to score combinations. Direct evaluation runs the assembled model for each combination it considers. Including precomputation, linearized search costs slightly more per seed at layer level and less at matrix level than direct evaluation (\cref{tab:search-total}).

\begin{table}[htb]
\centering\small
\setlength{\tabcolsep}{3pt}
\begin{tabular}{llcc}
\toprule
Units & Objective & Search time & \shortstack{Total cost\\(GPU-hours)} \\
\midrule
Layer & Linearized & 3--10 s & $\approx 2.6$ \\
      & Actual & 0.5--1.1 h & 2.22 \\
Matrix & Linearized & 14--69 s & $\approx 5.1$ \\
       & Actual & 3.7--5.4 h & 6.33 \\
\bottomrule
\end{tabular}
\caption{\textbf{Search time and total cost per seed.} Search times are wall-clock ranges on shared GPUs. Total costs include candidate preparation. Direct-evaluation costs are means over five seeds. Linearized costs are approximate, with timings available for four layer-level seeds and three matrix-level seeds.}
\label{tab:search-cost}\label{tab:search-total}
\end{table}

\section{Storage and deployment}\label{app:storage}\label{app:deployment-details}\label{sec:exp-eff}\label{app:txt-serve}
\paragraph{Deployment storage.}
At the same budget $K$, shared grouping, BRR-layer and BRR-matrix each store $K$ backbones' worth of merged weights to serve all 100 tasks. Finer grouping increases only the routing table, which stores one byte per task for each distinct grouping and occupies at most 7.3\,KB (\cref{tab:txt-storage}).

\begin{table*}[t]
\centering\small
\setlength{\tabcolsep}{3pt}
\small
\begin{tabular}{lccc}
\toprule
Deployment & Merged weights (backbones) & routing & total \\
\midrule
grouping-only operator, any grouping, $K$ merged units & $K$ & ${\le}\,7.3$\,KB & $K$ \\
FlexMerge official, $K{=}10$ / $K{=}5$ budget & 9.95 / 4.96 & 5.2\,KB & 9.95 / 4.96 \\
EMR or TALL masks, $K{=}10$ & 10 (+1 base) & ${\le}\,7.3$\,KB & 14.13 \\
100 experts & 100 & -- & 100 \\
\bottomrule
\end{tabular}
\caption{\textbf{Deployment storage for 100 ViT-B/32 tasks.} Merged weights and total storage are measured in backbone equivalents (0.326\,GiB each in FP32). Routing indices are included in the totals but are too small to affect the displayed values. The mask-based deployment adds 4.13 backbone equivalents: one for the base weights and 3.13 for the packed masks and, for EMR, its rescaling factors. FlexMerge stops at the first merge that brings storage below its target budget.}
\label{tab:txt-storage}
\end{table*}

\paragraph{Routed inference.}
Shared grouping, BRR-layer, BRR-matrix and FlexMerge select one stored module per sample in each unit, so each requires the same backbone computation as a single model. Our implementation processes the samples that use the same module together and reorders their activations whenever the grouping changes. A ViT-B/32 forward pass requires one such reordering for shared grouping, 13 for BRR-layer, 73 for BRR-matrix and 25 for FlexMerge, which applies one grouping to all matrices of an attention or MLP block. Channel Merging clusters each output channel separately, so a task's weight matrix is not one of the stored matrices. We serve it by gathering, for each matrix and each task in the batch, that task's rows from the stored copies and applying the gathered matrix to the task's samples. Computing all ten cluster outputs and then selecting one per channel was 2.4--4.2 times slower. For every method, on three tasks, the maximum absolute difference in pooled features between routed inference and a model assembled for the individual task was at most $9.1\times10^{-6}$.

We timed all five methods in a single run on ViT-B/32 with 100 tasks, $K=10$ and $\alpha=1$. The grouping methods use TSV-M merges, and Channel Merging uses its own merge rule; latency does not depend on the merged values. Measurements used one NVIDIA RTX 6000 Ada GPU that ran no other process, with tasks sampled uniformly and identical batches for all methods.

At batch sizes 64 and 256, BRR-layer added 2\% latency relative to shared routing, FlexMerge 6\%, BRR-matrix 14\%--16\% and Channel Merging 81\%--142\% (\cref{tab:txt-serve}). At batch size 16, where the reorderings are amortized over fewer samples, these overheads were 22\%, 28\%, 82\% and 93\%. FlexMerge's authors instead serve each task through its own view of the merged weights and process the tasks' sub-batches in parallel. Running these sub-batches one after another took 1.96--3.51 times as long as shared routing when tasks were drawn uniformly from all 100, which bounds the cost of parallel execution from above. This serving style runs one assembled model per task, so its cost is the same for shared grouping, BRR and FlexMerge. At the validation-selected scale, BRR-layer retains about 82\% of BRR-matrix's accuracy gain over shared grouping in this TSV-M setting (computed from unrounded gains; \cref{tab:main}), with much lower routing overhead at batch sizes 64 and 256.

\begin{table}[h]
\centering\small
\setlength{\tabcolsep}{4pt}
\begin{tabular}{llcc}
\toprule
Batch & Method & Median (ms) & vs.\ dense \\
\midrule
16 & Channel Merging & 28.6 & 3.55 \\
16 & FlexMerge & 18.9 & 2.34 \\
16 & shared & 14.8 & 1.83 \\
16 & BRR-layer & 18.0 & 2.23 \\
16 & BRR-matrix & 26.9 & 3.34 \\
\midrule
64 & Channel Merging & 91.9 & 3.10 \\
64 & FlexMerge & 40.4 & 1.36 \\
64 & shared & 38.0 & 1.28 \\
64 & BRR-layer & 38.7 & 1.31 \\
64 & BRR-matrix & 44.2 & 1.49 \\
\midrule
256 & Channel Merging & 250.3 & 2.11 \\
256 & FlexMerge & 147.1 & 1.24 \\
256 & shared & 138.4 & 1.17 \\
256 & BRR-layer & 140.6 & 1.19 \\
256 & BRR-matrix & 158.0 & 1.33 \\
\bottomrule
\end{tabular}
\caption{\textbf{Routed inference latency.} Median latency and its ratio to a single assembled model (``dense''; 8.1, 29.7 and 118.5\,ms at batch sizes 16, 64 and 256), with tasks sampled uniformly. Each of three passes tested all methods and batch sizes in random order, with five warm-up and 20 timed forward passes per condition and seed. We report the median of the 60 timed forward passes for each seed, averaged over three grouping seeds (Channel Merging: three clustering seeds; FlexMerge: its single deterministic grouping). Channel Merging gathers each task's rows per matrix; all other methods route samples to stored modules.}
\label{tab:txt-serve}
\end{table}

\section{Other experts and tasks}\label{app:other}
\subsection{Vision-language experiment details}
\label{app:txt-vlm}\label{sec:exp-vlm}

\subsubsection{Experts, tasks and metrics}
\label{app:txt-vlm-setup}

We use six visual question-answering (VQA) tasks---ScienceQA-IMG, IconQA, VSR, TallyQA, PathVQA and VizWiz---and four image-classification tasks presented as four-way multiple choice: Stanford Cars, FGVC-Aircraft, DTD and RESISC45. We score VizWiz by VQA accuracy, the other VQA tasks by exact or relaxed match, and the multiple-choice tasks by classification accuracy. The overall score is the unweighted mean of the ten task scores.

For each task, we fine-tune InternVL2.5-1B~\cite{chen2024internvl25} and Qwen2-VL-7B~\cite{wang2024qwen2vl} separately, updating all language-model parameters while keeping the vision tower and projector frozen. Both models train for 1,000 steps with batch size 16, AdamW, weight decay 0.01, 3\% warmup and a cosine learning-rate schedule. Learning rates selected on three pilot tasks are $10^{-5}$ for InternVL and $5\times10^{-6}$ for Qwen.

\subsubsection{Merge scale and uncertainty}
\label{app:txt-vlm-alpha}

For FlexMerge with TA, \cref{tab:vlm-fft,tab:vlm-k5} report the stronger variant, which scales the mean task vector in each group. The task-type breakdown in \cref{tab:vlm-groups} uses the sum of the task vectors for this baseline. Shared grouping, both BRR variants and the single merged model use group-sum scaling with TA throughout.

We select the merge scale on validation data for each grouping seed. All four merging methods start from the grid $\{0.1,0.3,0.5,0.7,1.0,1.5\}$. When an endpoint gives the highest validation score, we extend the grid in that direction. The search stops at 0.02 on the lower end and 4.0 on the upper end. Ties are resolved in favor of the smaller scale. In 23 Iso-CTS runs, validation selected the final upper bound of 4.0. For randomized methods, we first average the ten task scores within each seed, then report the mean and sample standard deviation across five seeds. FlexMerge uses one grouping.

\subsection{Additional vision-language results}\label{app:vlm-more}

\subsubsection{Results at $K=5$}\label{app:vlm-k5}

With ten experts and five balanced groups, each expert has just one partner per unit. On InternVL2.5-1B, mean scores still follow BRR-matrix $>$ BRR-layer $>$ shared grouping for all four merging methods (\cref{tab:vlm-k5}). On Qwen2-VL-7B, the three schemes are within 0.4 points of one another for each method, and the direction of the BRR-matrix gain over shared grouping depends on the method.
\begin{table*}[htb]
\centering\small
\setlength{\tabcolsep}{4pt}
\begin{tabular}{lcccccccc}
\toprule
 & \multicolumn{4}{c}{InternVL2.5-1B} & \multicolumn{4}{c}{Qwen2-VL-7B} \\
\cmidrule(lr){2-5}\cmidrule(lr){6-9}
Method & TA & TSV-M & Iso-CTS & ACE & TA & TSV-M & Iso-CTS & ACE \\
\midrule
Channel Merging & \multicolumn{4}{c}{$78.66_{\pm 0.05}$} & \multicolumn{4}{c}{$87.10_{\pm 0.07}$} \\
FlexMerge & $75.76$ & $79.05$ & $78.16$ & $80.13$ & $86.40$ & $86.91$ & $\mathbf{85.64}$ & $87.15$ \\
Shared & $79.27_{\pm 0.08}$ & $80.04_{\pm 0.21}$ & $78.37_{\pm 0.26}$ & $80.76_{\pm 0.08}$ & $86.71_{\pm 0.32}$ & $87.00_{\pm 0.18}$ & $84.47_{\pm 0.14}$ & $\mathbf{87.34}_{\pm 0.05}$ \\
BRR-layer & $79.88_{\pm 0.25}$ & $80.21_{\pm 0.20}$ & $78.46_{\pm 0.19}$ & $80.94_{\pm 0.06}$ & $87.05_{\pm 0.06}$ & $\mathbf{87.07}_{\pm 0.10}$ & $84.50_{\pm 0.08}$ & $87.23_{\pm 0.21}$ \\
BRR-matrix & $\mathbf{80.16}_{\pm 0.25}$ & $\mathbf{80.29}_{\pm 0.24}$ & $\mathbf{78.64}_{\pm 0.06}$ & $\mathbf{81.04}_{\pm 0.04}$ & $\mathbf{87.07}_{\pm 0.05}$ & $87.05_{\pm 0.11}$ & $84.58_{\pm 0.06}$ & $87.18_{\pm 0.13}$ \\
\bottomrule
\end{tabular}
\caption{\textbf{Vision-language results with two experts per balanced group ($K=5$).} Scores are ten-task macro-averages (\%). Randomized methods report the mean and sample standard deviation over five seeds. FlexMerge uses one grouping and group-mean scaling for TA. Channel Merging uses its own merge rule and yields one score per backbone. Bold marks the highest mean among FlexMerge, Shared, BRR-layer and BRR-matrix in each column.}
\label{tab:vlm-k5}
\end{table*}

\subsubsection{Question-answering and multiple-choice tasks}\label{app:vlm-groups}

\Cref{tab:vlm-groups} separates the two task types to check whether an overall gain conceals a decline in either category. On InternVL2.5-1B, BRR-matrix scores higher than shared grouping in both category means at both budgets for all four merging methods. On Qwen2-VL-7B, the result depends on the method. With ACE-Merging at $K=5$, for example, both category means are slightly lower under BRR-matrix.
\begin{table*}[htb]
\centering\footnotesize
\setlength{\tabcolsep}{4pt}
\begin{tabular}{llcccccccc}
\toprule
 & & \multicolumn{4}{c}{$K=2$} & \multicolumn{4}{c}{$K=5$} \\
\cmidrule(lr){3-6}\cmidrule(lr){7-10}
 & & \multicolumn{2}{c}{InternVL2.5-1B} & \multicolumn{2}{c}{Qwen2-VL-7B} & \multicolumn{2}{c}{InternVL2.5-1B} & \multicolumn{2}{c}{Qwen2-VL-7B} \\
Merging & Grouping & VQA & MC & VQA & MC & VQA & MC & VQA & MC \\
\midrule
TA & FlexMerge & 62.37 & 76.74 & 71.74 & 89.94 & 68.20 & 79.05 & 79.47 & 94.03 \\
 & Shared & 65.25 & 80.12 & 74.06 & 92.11 & 74.68 & 86.16 & 81.16 & 95.03 \\
 & BRR-layer & 66.47 & 80.15 & 76.14 & 93.66 & 75.46 & 86.49 & 81.70 & 95.08 \\
 & BRR-matrix & 66.60 & 80.21 & 76.27 & 93.63 & 75.87 & 86.60 & 81.74 & 95.06 \\
\midrule
TSV-M & FlexMerge & 72.09 & 83.99 & 80.79 & 94.97 & 75.34 & 84.61 & 81.66 & 94.79 \\
 & Shared & 72.81 & 85.12 & 80.38 & 95.05 & 75.64 & 86.64 & 81.60 & 95.10 \\
 & BRR-layer & 73.53 & 85.47 & 80.54 & 95.17 & 75.89 & 86.69 & 81.65 & 95.19 \\
 & BRR-matrix & 73.69 & 85.48 & 80.71 & 95.12 & 75.99 & 86.75 & 81.63 & 95.17 \\
\midrule
Iso-CTS & FlexMerge & 72.54 & 84.12 & 76.78 & 94.80 & 74.77 & 83.25 & 80.24 & 93.74 \\
 & Shared & 72.49 & 84.50 & 77.34 & 94.61 & 73.72 & 85.33 & 77.71 & 94.62 \\
 & BRR-layer & 72.85 & 84.75 & 77.61 & 94.68 & 73.86 & 85.37 & 77.71 & 94.68 \\
 & BRR-matrix & 73.04 & 84.54 & 77.71 & 94.64 & 74.10 & 85.44 & 77.84 & 94.69 \\
\midrule
ACE & FlexMerge & 74.88 & 84.80 & 81.59 & 95.06 & 76.39 & 85.73 & 81.81 & 95.16 \\
 & Shared & 74.75 & 85.89 & 81.76 & 95.11 & 76.64 & 86.96 & 82.14 & 95.14 \\
 & BRR-layer & 75.31 & 86.00 & 81.76 & 95.05 & 76.90 & 87.01 & 81.98 & 95.09 \\
 & BRR-matrix & 75.50 & 86.13 & 81.74 & 95.10 & 76.95 & 87.18 & 81.92 & 95.06 \\
\bottomrule
\end{tabular}

\caption{\textbf{Vision-language scores by task type at $K=2$ and $K=5$.} VQA is the mean score over six question-answering tasks, and MC is the mean accuracy over four multiple-choice tasks. Scores (\%) are averaged over seeds for randomized methods. FlexMerge with TA uses group-sum scaling in this table.}
\label{tab:vlm-groups}
\end{table*}

\nocite{chung2024flant5}%

\end{document}